\documentclass{article}

\PassOptionsToPackage{numbers, sort&compress}{natbib}

\usepackage[main, final]{neurips_2026}
\usepackage[utf8]{inputenc} % allow utf-8 input
\usepackage[T1]{fontenc}    % use 8-bit T1 fonts
\usepackage{hyperref}       % hyperlinks
\hypersetup{hidelinks}
\usepackage{url}            % simple URL typesetting
\usepackage{booktabs}       % professional-quality tables
\usepackage{amsfonts}       % blackboard math symbols
\usepackage{nicefrac}       % compact symbols for 1/2, etc.
\usepackage{microtype}      % microtypography
\usepackage{xcolor}         % colors
\usepackage{graphicx}
\usepackage{multirow}
\usepackage{amsmath}
\usepackage{float}
\usepackage{placeins}
\usepackage{caption}
\usepackage{wrapfig}
\usepackage[table]{xcolor}

\title{Less Supervision, Better Generalization: \\Weakly Supervised Fake Region Localization \\in Diffusion-Edited Images}

\author{%
  Junhee Lee \\
  Kyung Hee University \\
  \texttt{jhlee39@khu.ac.kr}
  \And
  Donghyeon Jeon \\
  Kyung Hee University \\
  \texttt{amuse\_dh@khu.ac.kr}
  \And
  Taeoh Kim \\
  NAVER Cloud \\
  \texttt{taeoh.kim@navercorp.com}
  \AND
  Beomyoung Kim \\
  NAVER Cloud \\
  \texttt{beomyoung.kim@navercorp.com}
  \And
  MyeongAh Cho\thanks{Corresponding author.} \\
  Kyung Hee University \\
  \texttt{maycho@khu.ac.kr}
  \AND
  \textnormal{\textbf{Project Page:} \url{https://visualsciencelab-khu.github.io/LSBG_project}}
}

\begin{document}

\maketitle

\begin{abstract}
Localizing AI-edited regions is essential for interpretable forensic analysis, but remains challenging due to subtle and spatially distributed artifacts that are misaligned with semantic or object boundaries.
Existing approaches rely on pixel-level supervision from controlled editing pipelines, which is difficult to scale and can introduce misleading signals: artifacts frequently extend beyond annotated regions, while out-of-mask pixels are treated as authentic. 
This limits models' ability to capture transferable evidence and generalize across generators and datasets.
To address these issues, we propose \textbf{ReGFLoW}, a \textbf{Re}construction-\textbf{G}uided \textbf{F}ake \textbf{Lo}calization framework under \textbf{W}eak supervision, which is the first weakly supervised approach for diffusion-edited fake region localization.
ReGFLoW requires only real/fake labels at the image-level and uses diffusion reconstruction errors as dense  spatial guidance to inject them into both feature and score spaces.
Furthermore, by artifact-centric multiple instance learning, ReGFLoW utilizes localized diffusion evidence without relying on semantic-affinity or boundary-based pseudo-mask priors.
Extensive experiments show competitive cross-generator localization, while ReGFLoW outperforms all evaluated fully supervised baselines when evaluation includes both partially edited and fully synthetic images and in cross-dataset tests, without target-domain adaptation.
\end{abstract}

\section{Introduction}
The rapid advancement of generative image models has significantly lowered the barrier to producing highly realistic visual content\citep{ho2020ddpm,rombach2022ldm,meng2022sdedit,avrahami2022blended,lugmayr2022repaint}.
While these developments enable a wide range of creative applications, they also raise serious concerns regarding the proliferation of synthesized or manipulated images online.
As a result, detecting AI-generated images has become a critical issue in digital forensics and security fields\citep{wang2023dire,koutlis2024rine,ricker2024aeroblade,wang2025opensdi}.
However, image-level detection alone provides limited forensic value, as it only indicates whether an image is suspicious without identifying the manipulated regions.
To address this limitation, AI-generated image region localization aims to determine authenticity at the regional level\citep{mareen2024tgif,bazyleva2025xedit,zhang2025deal300k,huang2025sida,mareen2026tgif2,cai2026brgen,costanzino2025radar,wang2025opensdi,kang2025legion}.
This enables more interpretable forensic analysis by providing explicit evidence of where manipulations have occurred, thereby improving explainability for real-world applications.

%%%%%%%%%%%%%%%%%%%%%%%%%%%%%%%%%%%%%%%%%%%%%%%%%%%%%%%%%%%%%%%%%%%%%
\begin{figure}[t]
    \centering
    \includegraphics[width=\linewidth]{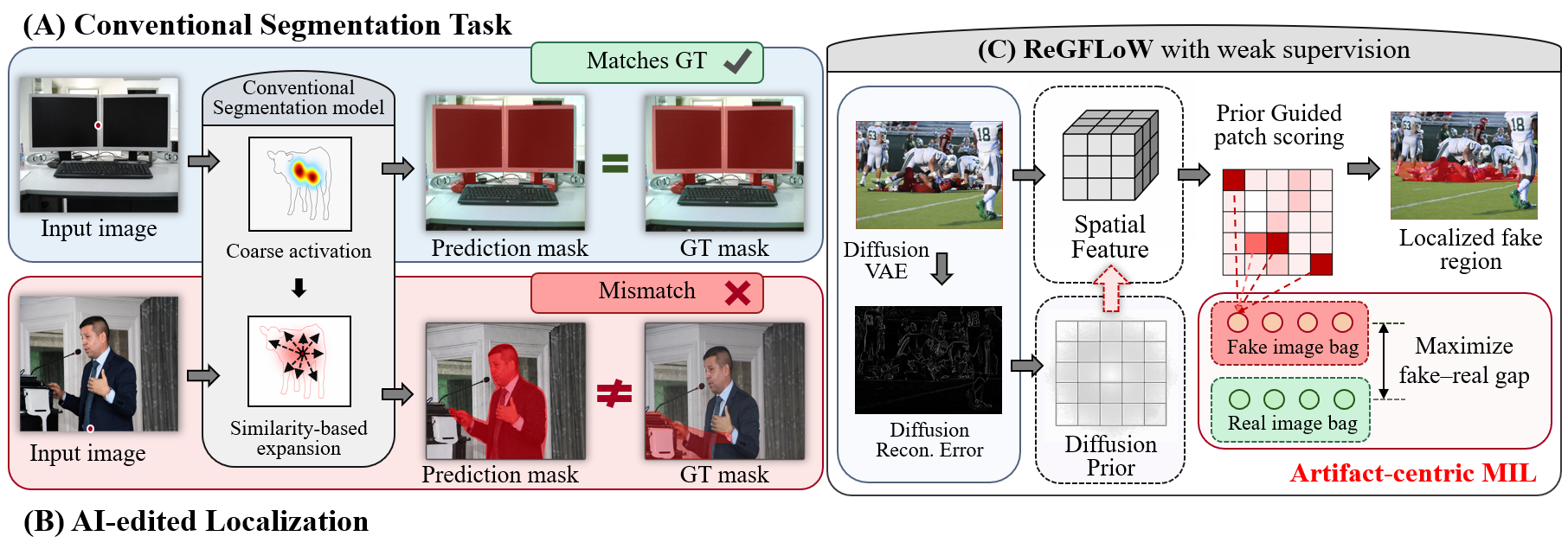}
    \caption{
Motivation of ReGFLoW.
(A) Conventional localization tasks, such as semantic segmentation and image manipulation localization, often rely on object- or boundary-aligned targets.
(B) Diffusion-edited fake regions can be non-semantic and weakly bounded, making similarity-based expansion unreliable.
(C) ReGFLoW introduces reconstruction-guided MIL for weakly supervised fake region localization.
    }
    \vspace{-1em}
    \label{fig:fig1}
\end{figure}
%%%%%%%%%%%%%%%%%%%%%%%%%%%%%%%%%%%%%%%%%%%%%%%%%%%%%%%%%%%%%%%%%%%%%

Among various generative image models, this paper focuses on localizing manipulated regions in images generated or edited by diffusion-based models.
Although this task appears similar to conventional semantic segmentation or image manipulation localization\citep{ma2023imlvit,guillaro2023trufor,guo2023hifi,zhai2023wscl,tantaru2024dolos}, it poses fundamentally different challenges that existing approaches cannot adequately address.
Semantic segmentation and manipulation localization methods are typically designed for objects or background regions with clear boundaries\citep{zhou2016cam,kolesnikov2016sec,ahn2018affinitynet,ahn2019irnet,wang2020seam,chen2022recam,ma2023imlvit,guillaro2023trufor,guo2023hifi,wu2019mantranet,kwon2021catnet,chen2021mvssnet}.
However, in diffusion-based editing, the changes are often less structured—for example, slight background alterations, partial edits to objects, or subtle texture inconsistencies that do not correspond to clear semantic regions.\cite{meng2022sdedit,avrahami2022blended,lugmayr2022repaint,mareen2024tgif,bazyleva2025xedit,zhang2025deal300k,huang2025sida,mareen2026tgif2,costanzino2025radar}
As a result, the diffusion-edited image localization task requires identifying low-level visual traces rather than relying on high-level semantic cues.

Existing methods for diffusion-edited region localization typically rely on fully supervised learning, using pixel-level manipulation masks generated during the image editing or inpainting process\citep{wang2025opensdi,huang2025sida,bazyleva2025xedit,zhang2025deal300k}.
While this approach is simple and effective in controlled benchmark scenarios, there are two reasons to argue that it is unsuitable for realistic diffusion-edited region localization.
\textbf{First}, obtaining accurate pixel-level ground-truth masks at scale is inherently challenging, like other segmentation tasks.
However, aside from this, commercial image editing and generation services mainly take text-based inputs, making it difficult to restrict edits to only the masked region.\citep{openai_chatgpt_images,adobe_firefly_masking,midjourney_editor,canva_magic_edit,capcut_ai_replace,google_imagen_mask_free,google_nano_banana,bfl_flux2_docs,bytedance_seedream4,qwen_image_edit,hunyuan_image3,omnigen}
Even if a mask image is provided or the target region is well specified in text, the model may still manipulate regions outside the intended region.
These issues make it difficult to obtain reliable edit masks during the image editing process.
\textbf{Second}, even if the mask is created or obtained by humans, it can lead to unintended behavior due to incorrect supervision.
As described above, diffusion-based image manipulation involves latent encoding, denoising, decoding, and blending processes that can leave generation traces beyond the intended editing region defined by the mask.\citep{rombach2022ldm,meng2022sdedit,avrahami2022blended,lugmayr2022repaint,ricker2024aeroblade,wang2023dire}
However, when imprecise mask-based fully supervised learning is applied without accounting for this, it treats all regions outside the mask as unmanipulated and enforces supervision strictly based on the mask boundaries.
Additionally, the misalignment between the mask and the manipulated regions may cause the model to incorrectly learn which regions are manipulated.
As a result, the model tends to rely on mask boundaries while ignoring subtle artifacts outside the mask region, leading to spurious correlations.
As a result, generalization performance in unseen domains is degraded, as analyzed in detail in Section~\ref{sec:rethinking}.

Motivated by these issues, we propose the first weakly supervised method for diffusion-edited fake region localization in generic visual content using only image-level real/fake labels.
This allows the model to learn from a wider range of edited images without relying on pixel-level masks.
It also avoids assuming that diffusion artifacts appear only in the edited regions, which helps improve generalization.
However, weakly supervised learning alone poses challenges for precise spatial part localization.
Existing weakly supervised approaches address conventional image manipulation~\cite{zhai2023wscl,guo2026boxpromptiml} or study diffusion edits in face-centric settings~\cite{tantaru2024dolos}. In our setting, diffusion-edited regions, as shown in Figure~\ref{fig:fig1}, are not consistently aligned with semantic object structures, limiting the effectiveness of conventional pseudo-mask refinement and requiring a new task-specific approach for diffusion-edited fake region localization~\cite{zhou2016cam,kolesnikov2016sec,ahn2018affinitynet,ahn2019irnet,wang2020seam,chen2022recam}.

In this paper, we propose \textbf{ReGFLoW}, \textbf{Re}construction-\textbf{G}uided \textbf{F}ake \textbf{Lo}calization under \textbf{W}eak supervision. 
ReGFLoW learns localization via an \emph{artifact-centric multiple instance learning} (MIL) objective, enabling the model to discover diffusion artifact evidence beyond the intended edit mask. 
The core idea of ReGFLoW is to overcome the limitations of semantic-affinity or boundary-based spatial priors by leveraging diffusion reconstruction error---previously studied mainly for image-level fake detection---as a prior to identifying fake regions.
Specifically, the diffusion reconstruction error serves as dense spatial guidance, injected into both the feature and score spaces.
This guidance allows the model to learn from the entire image by capturing generalizable diffusion artifact cues rather than relying solely on source-specific and mask-bound patterns.

Our contributions are summarized as follows:
\begin{itemize}
    \item We provide a systematic analysis of pixel-level mask supervision for diffusion-edited localization, identifying two generalization-limiting factors: pixel-level masks are often unavailable in realistic data, and mask-based supervised training can suppress transferable diffusion traces by treating out-of-mask regions as belonging to the real class.

    \item To overcome these generalization limits, we formulate, to the best of our knowledge, the first weakly supervised setting for diffusion-edited fake region localization in generic visual content, where models learn spatial localization from image-level real/fake labels rather than pixel-level manipulation masks.

    \item We propose \textbf{ReGFLoW}, a \textbf{Re}construction-\textbf{G}uided \textbf{F}ake \textbf{Lo}calization framework under \textbf{W}eak supervision. ReGFLoW repurposes diffusion reconstruction error from image-level fake detection into dense spatial guidance, and combines it with an artifact-centric multiple instance learning objective to localize diffusion artifacts without object- or boundary-based pseudo-mask priors.

    \item ReGFLoW achieves competitive cross-generator localization and outperforms all evaluated fully supervised baselines when evaluated on both partially edited and fully synthetic images and across datasets, without target-domain adaptation.
\end{itemize}

\section{Rethinking Supervision for Diffusion-Edited Localization}
\label{sec:rethinking}

\subsection{Pixel Masks Limit Broad-Domain Training}
\label{sec:mask_unavailable}
A central goal of diffusion-edited image detection is to build a detector that generalizes across diverse generators (including both models and services)~\cite{mareen2024tgif,bazyleva2025xedit,zhang2025deal300k,huang2025sida,mareen2026tgif2,costanzino2025radar,wang2025opensdi,kang2025legion} and image distributions.
Existing research methods train on a single source domain—such as a specific generator or image distribution—and evaluate on unseen domains, which limits cross-domain generalization since the training domain remains fixed.
However, real-world systems must be trained across a broader set of domains, as generative models and editing services continue to diversify rapidly~\cite{openai_chatgpt_images,midjourney_editor,canva_magic_edit,capcut_ai_replace,google_nano_banana,bfl_flux2_docs,bytedance_seedream4,qwen_image_edit,hunyuan_image3,omnigen}.
Therefore, it is important to move beyond such constrained protocols and more deeply study generalization for diffusion-edited image detection.

This goal exposes a fundamental limitation of pixel-level supervision, as localization requires pixel-level manipulation masks.
However, many commercial generative and editing systems are text-conditioned and do not provide mask-based interfaces, making such masks difficult to obtain from the final outputs (More details in Sec~\ref{app:mask_availability}).
As a result, pixel-level supervision restricts training to a limited set of controlled sources where masks are explicitly available.
Weak supervision is therefore particularly suitable for diffusion-edited localization: it not only reduces annotation cost by using image-level labels, but also enables learning from broader and more realistic data distributions (Results in Table~\ref{tab:broad_domain_training}).

\subsection{Pixel Masks Can Provide Misleading Supervision for Diffusion Artifacts}
\label{sec:mask_incomplete}

%%%%%%%%%%%%%%%%%%%%%%%%%%%%%%%%%%%%%%%%%%%%%%%%%%%%%%%%%%%%%%%%%%%%%
\begin{figure}[t]
    \centering
    \includegraphics[width=1.0\linewidth]{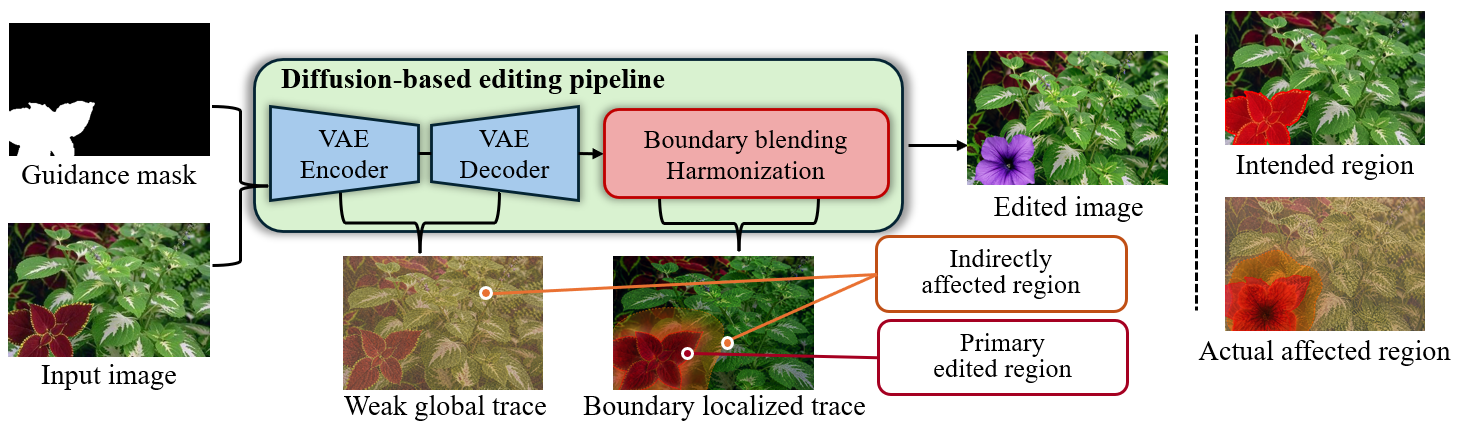}
    \caption{
    Conceptual illustration of mask incompleteness in diffusion-based editing. 
    The guidance mask indicates the intended edit region, but latent processing and blending can introduce weak global or boundary-localized traces beyond the mask. 
    Thus, the manipulation mask should not be treated as a complete label for all diffusion artifacts.
    }
    \vspace{-1em}
    \label{fig:fig2}
\end{figure}
%%%%%%%%%%%%%%%%%%%%%%%%%%%%%%%%%%%%%%%%%%%%%%%%%%%%%%%%%%%%%%%%%%%%%

Even when pixel-level masks are available, they can provide misleading supervision for learning diffusion artifacts.
While conventional image manipulations are often approximated with variations confined to selected regions~\citep{ma2023imlvit,guillaro2023trufor,guo2023hifi,wu2019mantranet,kwon2021catnet,chen2021mvssnet}, diffusion-based editing can affect a larger spatial region than the intended editing mask~\citep{rombach2022ldm,meng2022sdedit,avrahami2022blended,lugmayr2022repaint}.
As shown in Figure~\ref{fig:fig2}, the provided mask indicates the area where editing is intended, but, due to the nature of the diffusion model, the diffusion trace can be extended beyond that mask through mapping to the latent space via the VAE and subsequent blending or harmonization~\citep{rombach2022ldm,avrahami2022blended,lugmayr2022repaint,ricker2024aeroblade}.
Therefore, the masks provided alone should not be considered as a correct answer label for the diffusion-edited fake region localization.

These discrepancies directly affect supervised learning, and consequently also affect generalization.
Pixel mask supervised learning assigns fake labels to pixels inside the mask and real labels to all pixels outside the mask.
If the mask outer region contains weak but transferable diffusion traces, the model is trained to suppress evidence that may arise during the diffusion process itself~\citep{ricker2024aeroblade,wang2023dire}.
Thus, the model relies on stronger but less transferable clues, such as the boundaries of objects confined to masks, specific generative models, or patterns in dataset distributions~\citep{mareen2024tgif,bazyleva2025xedit,zhang2025deal300k,huang2025sida,mareen2026tgif2,wang2025opensdi}.
As a result, if these source-specific clues are absent or weakened in unseen regions, the affected regions may be misclassified as authentic rather than fake.

\begin{wrapfigure}{r}{0.5\columnwidth}
\vspace{-0.9\intextsep}
\centering
\includegraphics[
    width=\linewidth,
    trim=0 0 0 10pt,
    clip
]{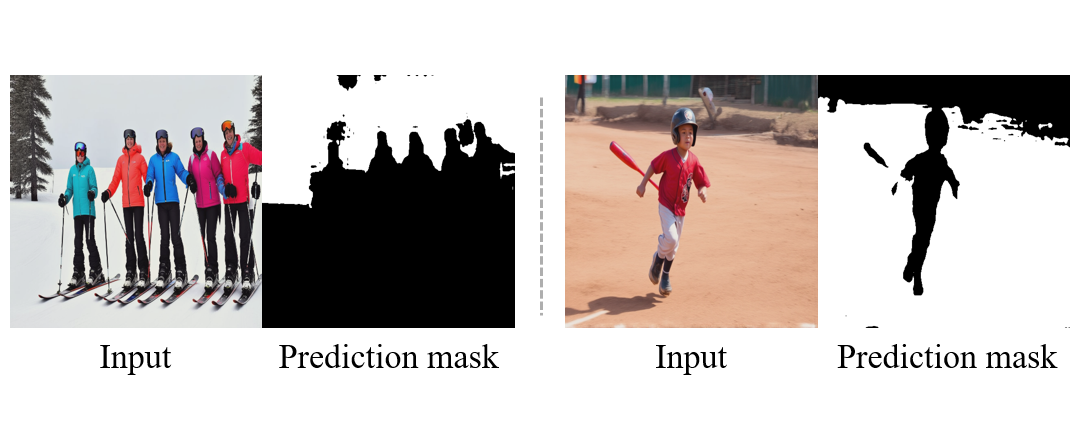}
\caption{
Failure case on an out-of-domain fully synthetic image.
Although the entire image is fake, the mask-supervised baseline predicts only an object-shaped region and misses large fake areas.
}
\label{fig:fully_synthetic_failure}
\vspace{-0.6\intextsep}
\end{wrapfigure}

We observe this tendency in a qualitative motivating example.
As shown in Figure~\ref{fig:fully_synthetic_failure}, the input is an out-of-domain fully synthetic image, where the entire image should be treated as fake.
However, the fully supervised baseline predicts only an object-shaped region and suppresses large fake areas as real.
This indicates that the model does not simply make boundary errors, but can fail to recognize fake evidence outside the mask-aligned patterns learned during training.
%%%%%%%%%%%%%%%%%%%%%%%%%%%%%%%%%%%%%%%%%%%%%%%%%%%%%%%%%%%%%%%%%%%%%
%%%%%%%%%%%%%%%%%%%%%%%%%%%%%%%%%%%%%%%%%%%%%%%%%%%%%%%%%%%%%%%%%%%%%
This behavior suggests that mask-supervised training can bias the model toward mask-aligned or source-specific cues, rather than transferable diffusion artifacts.
When such cues do not hold in unseen domains, fake regions can be missed and classified as real.
These observations motivate moving beyond a purely fully supervised mask-prediction paradigm.
Rather than serving only as a lower-cost alternative to dense annotation, weak supervision provides a better-aligned formulation for learning generalizable diffusion artifact cues from image-level labels without forcing all out-of-mask regions to be labeled as real
\citep{dietterich1997mil,ilse2018attentionmil,pinheiro2015imagelevel,zhai2023wscl,tantaru2024dolos}.

\vspace{-0.5em}
\section{Method}
\label{sec:method}

%%%%%%%%%%%%%%%%%%%%%%%%%%%%%%%%%%%%%%%%%%%%%%%%%%%%%%%%%%%%%%%%%%%%%
\begin{figure}[t]
    \centering
    \includegraphics[width=\linewidth]{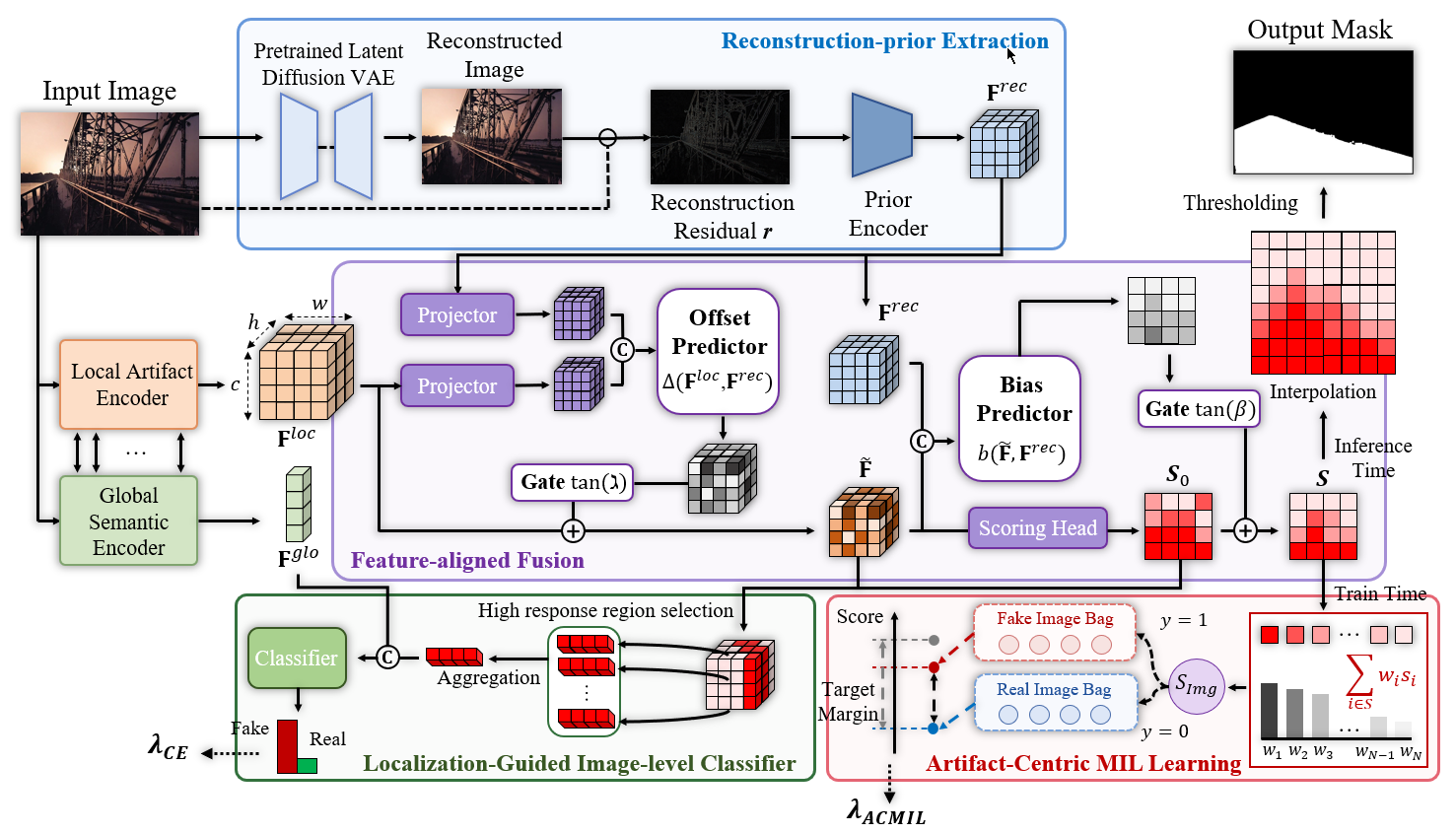}
    \caption{
Overview of the proposed ReGFLoW framework.
ReGFLoW integrates dual-encoder features and a diffusion reconstruction prior for patch-level fake evidence estimation under weak supervision.
The resulting score map is optimized with artifact-centric MIL and used for fake-region localization and image-level classification.
    }
    \vspace{-1em}
    \label{fig:fig3}
\end{figure}
%%%%%%%%%%%%%%%%%%%%%%%%%%%%%%%%%%%%%%%%%%%%%%%%%%%%%%%%%%%%%%%%%%%%%

\vspace{-1em}
\subsection{Overview}

Figure~\ref{fig:fig3} illustrates the overall pipeline.
ReGFLoW learns fake region localization in diffusion-edited images using only image-level real/fake labels.
Since diffusion-edited artifacts include both semantic inconsistencies and
subtle low-level traces, we use a dual-encoder backbone to capture global
semantic cues and local artifact-sensitive features, with intermediate
cross-attention exchanging information between the two streams\citep{radford2021clip,he2022mae}.

Starting from the resulting dense feature representation, the model proceeds
in three steps:
(i) a diffusion reconstruction error map is computed and aligned with the
local feature space to provide a dense patch-level prior,
(ii) the prior is injected into the feature and score spaces through a
\emph{reconstruction-guided} patch scoring module, producing a patch score map,
and
(iii) the score map is trained end-to-end with an artifact-centric
multiple instance learning (MIL) objective using only image-level labels\citep{dietterich1997mil,sultani2018real,pinheiro2015imagelevel}.
In addition, the localization feature and score map guide the image-level
classifier by providing artifact-focused regional evidence.

\subsection{Reconstruction-Guided Patch Scoring}
\label{sec:recon}
\vspace{-1em}
\paragraph{Diffusion reconstruction error as a dense prior.}
Reconstruction error from pretrained diffusion autoencoders has been used as an effective cue for AI-generated image detection~\citep{wang2023dire,ricker2024aeroblade}.
Following this observation, we use the frozen Variational Autoencoder (VAE) of
SD1~\citep{compvis2022stableDiffusionV11,rombach2022ldm} as a diffusion-aligned reconstructor,
since it provides a widely used latent-diffusion reconstruction prior without
additional training.
Since diffusion-synthesized content has already been projected through this
learned latent manifold, it is often reconstructed with a smaller or more
regular residual than natural real-image content.
We adapt this reconstruction residual as a spatial prior for localization,
allowing the network to relate reconstruction patterns to diffusion artifacts at each image region.

Given an input image $I$, we compute the reconstruction residual using the
frozen VAE of SD1:
\begin{equation}
    \mathbf{r} \;=\; \bigl|\,I - \mathcal{D}\!\bigl(\mathcal{E}(I)\bigr)\bigr|,
    \label{eq:recon}
\end{equation}
where $\mathcal{E}$ and $\mathcal{D}$ are the frozen VAE encoder and decoder,
respectively, and the absolute difference is taken channel-wise.
We use $\mathbf{r}$ as the reconstruction prior.

\paragraph{Feature-aligned fusion.}
Let $\operatorname{Enc}_{\mathrm{loc}}$ denote the local artifact encoder, which
produces a $32\times32$ patch feature map from the input image.
We obtain
$\mathbf{F}^{\mathrm{loc}}=\operatorname{AvgPool}_{2\times}(\operatorname{Enc}_{\mathrm{loc}}(I))
\in\mathbb{R}^{B\times C\times16\times16}$, where $C=768$.
This compact grid prevents the MIL bag from becoming too large, which improves
loss propagation and stabilizes weakly supervised training.
Since $\mathbf{r}$ is defined in image space, we align it to the same
$16\times16$ patch grid before fusion.
Let $\operatorname{Enc}_{\mathrm{prior}}$ denote the three-layer convolutional
Prior Encoder; we compute
$\mathbf{F}^{\mathrm{rec}}=\operatorname{AvgPool}_{2\times}(\operatorname{Enc}_{\mathrm{prior}}(\mathbf{r}))
\in\mathbb{R}^{B\times C_{\mathrm{rec}}\times16\times16}$, where
$C_{\mathrm{rec}}=32$.
We inject the reconstruction prior in two complementary ways:
(i) Feature-aligned Fusion through residual feature correction and
(ii) Bias Prediction through a zero-initialized reconstruction logit bias.

\paragraph{Residual feature fusion.}
Let $P_{\mathrm{loc}}$ and $P_{\mathrm{rec}}$ denote two separate convolutional
Projectors for $\mathbf{F}^{\mathrm{loc}}$ and $\mathbf{F}^{\mathrm{rec}}$,
respectively.
The convolutional Offset Predictor $G_{\Delta}$ then predicts the correction
offset from their channel-wise concatenation:
\begin{equation}
    \Delta(\mathbf{F}^{\mathrm{loc}},\mathbf{F}^{\mathrm{rec}})
    =
    G_{\Delta}\!\left(
    \left[
    P_{\mathrm{loc}}(\mathbf{F}^{\mathrm{loc}});
    P_{\mathrm{rec}}(\mathbf{F}^{\mathrm{rec}})
    \right]\right).
    \label{eq:delta}
\end{equation}
The offset provides a gated residual correction, so reconstruction cues can
adjust local artifact features without replacing them:
\begin{equation}
    \widetilde{\mathbf{F}}
    =
    \mathbf{F}^{\mathrm{loc}}
    +
    \tanh(\gamma)\cdot
    \Delta(\mathbf{F}^{\mathrm{loc}},\,\mathbf{F}^{\mathrm{rec}}),
    \label{eq:fuse}
\end{equation}
where $\gamma$ is a learnable scalar initialized to $0.05$.
The $\tanh(\cdot)$ gate makes the fusion start as a near-identity operation
and gradually activates as training progresses, preventing the reconstruction
prior from destabilizing early-stage optimization.
The resulting feature
$\widetilde{\mathbf{F}}\in\mathbb{R}^{B\times C\times16\times16}$ is used for
patch-level scoring.

\paragraph{Zero-initialized reconstruction logit bias.}
The fused feature $\widetilde{\mathbf{F}}$ is passed to the one-hidden-layer
convolutional Scoring Head, which produces a base patch logit map
$\mathbf{S}_0\in\mathbb{R}^{B\times16\times16}$.
To further refine the final patch scores, the Bias Predictor $b$ takes the
channel-wise concatenation of $\widetilde{\mathbf{F}}$ and
$\mathbf{F}^{\mathrm{rec}}$ and predicts a reconstruction logit bias
$b(\widetilde{\mathbf{F}},\mathbf{F}^{\mathrm{rec}})
\in\mathbb{R}^{B\times16\times16}$:
\begin{equation}
    \mathbf{S}
    \;=\;
    \mathbf{S}_0
    \;+\;
    \tanh(\beta)\cdot
    b(\widetilde{\mathbf{F}},\,\mathbf{F}^{\mathrm{rec}}),
    \label{eq:bias}
\end{equation}
where the final $1\times1$ output convolution of $b$ is
\emph{zero-initialized}, and $\beta$ is a learnable scalar initialized to $0.05$.
Because the bias branch initially outputs zero, the model starts with
$\mathbf{S}=\mathbf{S}_0$, i.e., it behaves identically to a baseline without reconstruction logit bias; the bias term grows only as training provides evidence for its utility.

Together, Eqs.\,\eqref{eq:delta}--\eqref{eq:bias} inject the reconstruction
prior into both feature and score spaces, while the gated residual path and
zero-initialized bias keep the initial behavior close to the baseline for stable
joint optimization.

\subsection{Weakly Supervised Training Objective}
\label{sec:objective}

\paragraph{MIL view of patch scoring.}
The final logit map
$\mathbf{S}\in\mathbb{R}^{B\times16\times16}$
is interpreted as a bag of patch instances.
For a fake image, at least some patches are expected to contain
diffusion-generated traces, whereas a real image should not contain such
artifact-positive patches.
Let $\mathcal{F}$ and $\mathcal{R}$ denote the fake and real image sets in a
batch, respectively.
The objective encourages the aggregated artifact score of each fake image to
exceed that of each real image, i.e.,
$\hat{s}_i > \hat{s}_j$ for $i\in\mathcal{F}$ and $j\in\mathcal{R}$, where
$\hat{s}$ denotes image-level artifact evidence aggregated from patch logits.

Compared with global pooling, which can dilute small fake regions or encourage broad responses, MIL focuses supervision on the most suspicious patches.
This better matches the partial and spatially variable nature of fake regions while still supporting fully synthetic images.

\paragraph{Artifact-centric MIL loss.}
Let $\{s_{i,n}\}_{n=1}^{N}$ denote the flattened patch logits of image $i$,
where $N=16\times16$.
Rather than using hard top-$k$ selection, we use image-wise normalized softmax
aggregation to handle large variations in fake-region size\citep{ilse2018attentionmil}:
\begin{equation}
    \hat{s}_i
    =
    \sum_{n=1}^{N}\alpha_{i,n}s_{i,n},
    \qquad
    \alpha_{i,n}
    =
    \frac{\exp(\kappa z_{i,n})}
    {\sum_{m=1}^{N}\exp(\kappa z_{i,m})},
    \qquad
    z_{i,n}
    =
    \frac{s_{i,n}-\mu_i}{\sigma_i+\epsilon},
    \label{eq:bag}
\end{equation}
where $z_{i,n}$ is the image-wise standardized logit, $\mu_i$ and $\sigma_i$
are the mean and standard deviation of the patch logits in image $i$, and
$\kappa$ controls the sharpness of the aggregation.
This aggregation acts as a soft top-$k$ operator, emphasizing patches with strong artifact evidence while remaining fully differentiable.
This is well suited to weakly supervised fake region localization in diffusion-edited images, where artifact traces may be weak and spatially diffuse, and the fake region size can vary across samples.

We enforce this ordering with a margin-based pairwise artifact-centric MIL loss:
\begin{equation}
    \mathcal{L}_{\mathrm{acmil}}
    \;=\;
    \frac{1}{|\mathcal{F}||\mathcal{R}|}
    \sum_{\substack{i\in\mathcal{F} \\ j\in\mathcal{R}}}
    \operatorname{softplus}\!\bigl(m - \hat{s}_i + \hat{s}_j\bigr),
    \label{eq:acmil}
\end{equation}
where $m>0$ is a margin hyperparameter.
This loss penalizes fake--real pairs whose artifact-score gap is smaller than the margin, thereby propagating image-level supervision to the patch score map.

\paragraph{Image-level branch.}
\label{sec:imglevel}
In addition to localization, ReGFLoW predicts an image-level real/fake label. The selected CLS features from the global semantic encoder are aggregated with a single-layer attention module to obtain a global image embedding.
Rather than classifying this embedding alone, we aggregate the top $10\%$
high-response locations of $\widetilde{\mathbf{F}}$ using masked softmax
attention from image-wise normalized $\mathbf{S}$.
The resulting artifact-focused context is concatenated with the global image
embedding and classified by a two-layer MLP with standard cross-entropy
$\mathcal{L}_{\mathrm{ce}}$.

\paragraph{Overall objective.}
The full training objective is
\begin{equation}
    \mathcal{L}
    =
    \lambda_{\mathrm{acmil}}\,\mathcal{L}_{\mathrm{acmil}}
    +
    \lambda_{\mathrm{ce}}\,\mathcal{L}_{\mathrm{ce}}
    +
    \lambda_{\mathrm{real}}\,\mathcal{L}_{\mathrm{real}}
    +
    \lambda_{\mathrm{smooth}}\,\mathcal{L}_{\mathrm{smooth}}.
    \label{eq:loss}
\end{equation}
Here, $\mathcal{L}_{\mathrm{real}}$ is a real-image hard-negative suppression
term that discourages high fake scores on real-image patches, and
$\mathcal{L}_{\mathrm{smooth}}$ is an edge-aware smoothness term that promotes
spatially coherent localization while preserving image boundaries.
Their detailed formulations are provided in Appendix~\ref{app:loss}.

\paragraph{Inference and calibration.}
At inference, $\mathbf{S}$ is bilinearly upsampled to the input resolution.
Before thresholding, we apply image-wise adaptive calibration to account for sample-dependent score distributions under weak supervision.
Each score map is normalized by its own mean and standard deviation, with the offset adjusted according to its mean uncalibrated positive response.
The calibrated probability map is then thresholded to obtain the final
fake-region mask.
Detailed formulations are provided in Appendix~\ref{app:calib}.

%%%%%%%%%%%%%%%%%%%%%%%%%%%%%%%%%%%%%%%%%%%%%%%%%%%%%%%%%%%%%%%%%%%%%

\begin{table*}[t]
\centering
\setlength{\abovecaptionskip}{2pt}
\setlength{\belowcaptionskip}{-3pt}
\caption{
P setting (partial edited images only). Pixel-level performance on OpenSDID benchmark. Existing weakly supervised baselines are adapted from conventional manipulation or face-centric deepfake localization. OOD Avg. denotes the average over four cross-domain generators: SD2.1, SDXL, SD3, and Flux.1.
}
\label{tab:opensdi_p}
\renewcommand{\arraystretch}{0.85}
\setlength{\tabcolsep}{3pt}
\footnotesize
\begin{tabular*}{\textwidth}{@{\extracolsep{\fill}}l*{14}{c}@{}}
\toprule
\multirow{3}{*}{\textbf{Method}}
& \multicolumn{2}{c}{\textbf{In-domain}}
& \multicolumn{8}{c}{\textbf{Cross-domain}}
& \multicolumn{4}{c}{\textbf{Average}} \\
\cmidrule(lr){2-3}
\cmidrule(lr){4-11}
\cmidrule(lr){12-15}
& \multicolumn{2}{c}{\textbf{SD1.5}}
& \multicolumn{2}{c}{\textbf{SD2.1}}
& \multicolumn{2}{c}{\textbf{SDXL}}
& \multicolumn{2}{c}{\textbf{SD3}}
& \multicolumn{2}{c}{\textbf{Flux.1}}
& \multicolumn{2}{c}{\textbf{Avg.}}
& \multicolumn{2}{c}{\textbf{OOD Avg.}} \\
\cmidrule(lr){2-3}
\cmidrule(lr){4-5}
\cmidrule(lr){6-7}
\cmidrule(lr){8-9}
\cmidrule(lr){10-11}
\cmidrule(lr){12-13}
\cmidrule(lr){14-15}
& \textbf{F1} & \textbf{IoU}
& \textbf{F1} & \textbf{IoU}
& \textbf{F1} & \textbf{IoU}
& \textbf{F1} & \textbf{IoU}
& \textbf{F1} & \textbf{IoU}
& \textbf{F1} & \textbf{IoU}
& \textbf{F1} & \textbf{IoU} \\
\midrule

\multicolumn{15}{@{}l}{\textit{Fully-supervised methods}} \\

PSCC-Net~\citep{liu2022psccnet}
& 64.2 & 54.7
& 44.8 & 36.7
& 26.1 & 19.7
& 37.3 & 29.3
& 11.6 & 8.2
& 36.8 & 29.7
& 29.9 & 23.5 \\

TruFor~\citep{guillaro2023trufor}
& 71.0 & 63.4
& 61.9 & 54.7
& 31.9 & 26.5
& 38.5 & 32.2
& 9.7 & 7.6
& 42.6 & 36.9
& 35.5 & 30.3 \\

IML-ViT~\citep{ma2023imlvit}
& 73.6 & 66.5
& 50.6 & 44.8
& 26.0 & 21.5
& 28.3 & 23.6
& 7.9 & 6.1
& 37.3 & 32.5
& 28.2 & 24.0 \\

MaskCLIP~\citep{wang2025opensdi}
& \textbf{75.8} & \textbf{68.6}
& \textbf{63.2} & \textbf{56.1}
& \textbf{35.2} & \textbf{29.5}
& \textbf{49.8} & \textbf{42.7}
& \textbf{18.4} & \textbf{14.8}
& \textbf{48.5} & \textbf{42.4}
& \textbf{41.7} & \textbf{35.8} \\

\midrule

\multicolumn{15}{@{}l}{\textit{Weakly-supervised methods}} \\

WSCL~\citep{zhai2023wscl}
& 25.6 & 16.7
& 25.9 & 17.0
& 26.0 & 17.0
& 25.4 & 16.5
& 25.3 & 16.4
& 25.6 & 16.7
& 25.7 & 16.7 \\

DOLOS~\citep{tantaru2024dolos}
& 21.5 & 13.4
& 19.1 & 11.9
& 19.8 & 12.2
& 20.3 & 12.5
& 19.2 & 12.0
& 20.0 & 12.4
& 19.6 & 12.2 \\

BoxPromptIML~\citep{guo2026boxpromptiml}
& 11.5 & 7.0
& 10.9 & 6.5
& 10.9 & 6.6
& 13.8 & 8.4
& 10.3 & 6.2
& 11.5 & 6.9
& 11.5 & 6.9 \\

\textbf{Ours}
& \textbf{49.3} & \textbf{36.9}
& \textbf{46.3} & \textbf{33.8}
& \textbf{34.9} & \textbf{23.6}
& \textbf{43.9} & \textbf{30.9}
& \textbf{29.2} & \textbf{19.0}
& \textbf{40.7} & \textbf{28.8}
& \textbf{38.6} & \textbf{26.8} \\

\bottomrule
\end{tabular*}
\end{table*}

%%%%%%%%%%%%%%%%%%%%%%%%%%%%%%%%%%%%%%%%%%%%%%%%%%%%%%%%%%%%%%%%%%%%%

\vspace{-0.2em}
\section{Experiments}
\label{sec:experiments}
\vspace{-0.5em}

\subsection{Experimental Setup}
\label{sec:exp_setup}
We conduct experiments on OpenSDID, a benchmark for diffusion-generated and
diffusion-edited image detection and localization~\citep{wang2025opensdi}.
Following the OpenSDID protocol, we train on the SD1.5 split and evaluate on
SD1.5 as the in-domain setting and on SD2.1, SDXL, SD3, and Flux.1 as
cross-domain settings.
Although OpenSDID provides both image-level real/fake labels and pixel-level
masks, ReGFLoW uses only image-level labels during training.
Pixel-level masks are reserved solely for evaluation.
As weakly supervised baselines, we adapt WSCL~\citep{zhai2023wscl} and BoxPromptIML~\citep{guo2026boxpromptiml} from conventional image manipulation localization, and DOLOS~\citep{tantaru2024dolos} from face-centric diffusion-edited localization.
We additionally evaluate cross-dataset localization on COCO-GLIDE~\citep{nichol2022glide} and DOLOS~\citep{tantaru2024dolos}, without target-domain adaptation.
Architectural and training details, including the backbone
configuration, are provided in Appendix~\ref{app:implementation}.

\subsection{Pixel-Level Localization Evaluation}
\label{sec:opensdi_localization}
Table~\ref{tab:opensdi_p}evaluates localization on partially edited fake images. The adapted weakly supervised baselines yield poor localization even in-domain, suggesting that their formulations do not readily extend to this task. Fully supervised baselines achieve strong in-domain scores on SD1.5 but degrade sharply on unseen generators. This is consistent with our analysis in Section~\ref{sec:rethinking} that pixel-mask supervision can encourage reliance on mask-confined, source-specific cues that transfer poorly across generators. In contrast, ReGFLoW, despite using only image-level labels, achieves competitive OOD Avg F1 (38.6), surpassing most fully supervised baselines with markedly more stable cross-domain behavior. The benefit is most pronounced on Flux.1 (29.2 vs. 18.4), the generator most distant from SD1.5, demonstrating that reconstruction-guided weak supervision effectively captures transferable diffusion traces where mask-bounded supervision fails.

Table~\ref{tab:opensdi_pf} evaluates the more realistic setting where both training and evaluation contain partially edited and fully synthetic fake images. Here the limitation of mask-based supervision becomes far more severe: MaskCLIP retains high in-domain F1 (95.1) but collapses across all unseen generators, yielding only 29.1 OOD Avg. ReGFLoW achieves the highest OOD Avg. F1 of 31.3, outperforming all evaluated fully supervised baselines while using no pixel-level supervision or target-domain adaptation. This suggests stronger generalization not only across image distributions, but also across fake-image types, as localization in this setting must handle both localized edits and fully generated images.

Beyond the cross-generator evaluation within OpenSDID, we further assess cross-dataset generalization under broader distribution shifts using COCO-GLIDE~\citep{guillaro2023trufor,nichol2022glide} and DOLOS~\citep{tantaru2024dolos}. Unlike the latent-space generative pipelines used in OpenSDID, GLIDE performs diffusion directly in image space, introducing a distinct generation mechanism, while DOLOS targets face-centric deepfakes, resulting in a substantial shift in content domain.
As shown in Table~\ref{tab:cross_dataset}, ReGFLoW achieves the highest F1 across all three settings, outperforming every evaluated fully supervised baseline without target-domain adaptation.

%%%%%%%%%%%%%%%%%%%%%%%%%%%%%%%%%%%%%%%%%%%%%%%%%%%%%%%%%%%%%%%%%%%%%
\begin{table*}[t]
\centering
\caption{
P+F setting (partial edited and fully synthetic images). Pixel-level F1 evaluated on OpenSDID. OOD Avg. denotes the average over four cross-domain generators.
}
\label{tab:opensdi_pf}
\small
\setlength{\tabcolsep}{4.6pt}
\renewcommand{\arraystretch}{0.8}
\begin{tabular*}{0.86\textwidth}{@{\extracolsep{\fill}}lccccccc@{}}
\toprule
\multirow{2}{*}{\textbf{Method}}
& \multicolumn{1}{c}{\textbf{In-domain}}
& \multicolumn{4}{c}{\textbf{Cross-domain}}
& \multicolumn{2}{c}{\textbf{Average}} \\
\cmidrule(lr){2-2}
\cmidrule(lr){3-6}
\cmidrule(lr){7-8}
& \textbf{SD1.5}
& \textbf{SD2.1}
& \textbf{SDXL}
& \textbf{SD3}
& \textbf{Flux.1}
& \textbf{Avg.}
& \textbf{OOD Avg.} \\
\midrule
\multicolumn{8}{@{}l}{\textit{Fully-supervised methods}} \\
TruFor~\citep{guillaro2023trufor}
& 28.8 & 20.1 & 15.9 & 16.4 & \textbf{8.4}
& 17.9 & 15.2 \\
IML-ViT~\citep{ma2023imlvit}
& 30.2 & 24.7 & 17.6 & 13.5 & 7.7
& 18.7 & 15.9 \\
MaskCLIP~\citep{wang2025opensdi}
& \textbf{95.1} & \textbf{73.2} & \textbf{20.8} & \textbf{17.7} & 4.8
& \textbf{42.3} & \textbf{29.1} \\
\midrule
\multicolumn{8}{@{}l}{\textit{Weakly-supervised methods}} \\
\textbf{Ours}
& \textbf{35.6} & \textbf{33.4} & \textbf{30.1} & \textbf{32.2} & \textbf{29.4}
& \textbf{32.2} & \textbf{31.3} \\
\bottomrule
\end{tabular*}
\end{table*}
%%%%%%%%%%%%%%%%%%%%%%%%%%%%%%%%%%%%%%%%%%%%%%%%%%%%%%%%%%%%%%%%%%%%%

%%%%%%%%%%%%%%%%%%%%%%%%%%%%%%%%%%%%%%%%%%%%%%%%%%%%%%%%%%%%%%%%%%%%%
\begin{table*}[t]
\centering
\caption{
Cross-dataset pixel-level F1 on COCO-GLIDE~\citep{nichol2022glide} and DOLOS~\citep{tantaru2024dolos}.
P denotes partially edited images, while P+F includes both partially edited and fully synthetic images.
All methods are evaluated without target-domain adaptation.
}
\label{tab:cross_dataset}

\small
\setlength{\tabcolsep}{4.6pt}
\renewcommand{\arraystretch}{0.85}

\begin{tabular*}{0.86\textwidth}{@{\extracolsep{\fill}}llccc@{}}
\toprule
\multirow{2}{*}{\textbf{Supervision}}
& \multirow{2}{*}{\textbf{Method}}
& \multicolumn{1}{c}{\textbf{COCO-GLIDE}}
& \multicolumn{2}{c}{\textbf{DOLOS}} \\
\cmidrule(lr){3-3}
\cmidrule(lr){4-5}
& &
\textbf{P}
& \textbf{P}
& \textbf{P+F} \\
\midrule

\multirow{3}{*}{\textit{Fully-supervised}}
& TruFor~\citep{guillaro2023trufor}
& \textbf{25.17} & 8.21 & 12.70 \\

& IML-ViT~\citep{ma2023imlvit}
& 15.23 & 1.16 & 1.74 \\

& MaskCLIP~\citep{wang2025opensdi}
& 8.34 & \textbf{13.35} & \textbf{14.65} \\

\midrule

\textit{Weakly-supervised}
& \textbf{Ours}
& \textbf{42.01} & \textbf{21.09} & \textbf{21.19} \\

\bottomrule
\end{tabular*}
\end{table*}
%%%%%%%%%%%%%%%%%%%%%%%%%%%%%%%%%%%%%%%%%%%%%%%%%%%%%%%%%%%%%%%%%%%%%

\subsection{Image-Level Detection Evaluation}
\label{sec:image_detection}
Table~\ref{tab:tab3} reports image-level detection performance on OpenSDID. Although ReGFLoW is primarily designed for localization, its image-level branch directly leverages localization information by fusing high-response regions from the patch score map with the global CLS embedding before classification, providing the detector with artifact-focused regional evidence rather than relying solely on a whole-image representation. As a result, ReGFLoW achieves the highest OOD Avg, outperforming not only image-only baselines but also methods that additionally use pixel-mask supervision. Notably, image-only baselines generally surpass methods jointly trained with pixel-mask supervision in OOD Avg, suggesting that mask annotations bias the model toward source-specific spatial patterns that hinder cross-domain transfer even at the image level. ReGFLoW circumvents this trade-off by deriving regional evidence from weakly supervised, reconstruction-guided localization, thereby benefiting from spatial information without inheriting the generalization cost of pixel-mask supervision.

%%%%%%%%%%%%%%%%%%%%%%%%%%%%%%%%%%%%%%%%%%%%%%%%%%%%%%%%%%%%%%%%%%%%%
\begin{table*}[t]
  \centering
  \caption{
  Image-level detection performance on OpenSDID. OOD Avg denotes the average over four cross-domain generators: SD2.1, SDXL, SD3, and Flux.1.
  }
  \label{tab:tab3}
  \setlength{\tabcolsep}{3pt}
  \renewcommand{\arraystretch}{1.08}
  \resizebox{\textwidth}{!}{%
  \begin{tabular}{@{}cl*{14}{c}@{}}
    \toprule
    \multirow{2}{*}{Approach} & \multirow{2}{*}{Method}
      & \multicolumn{2}{c}{SD1.5}
      & \multicolumn{2}{c}{SD2.1}
      & \multicolumn{2}{c}{SDXL}
      & \multicolumn{2}{c}{SD3}
      & \multicolumn{2}{c}{Flux.1}
      & \multicolumn{2}{c}{Avg}
      & \multicolumn{2}{c}{OOD Avg} \\
    \cmidrule(lr){3-4}
    \cmidrule(lr){5-6}
    \cmidrule(lr){7-8}
    \cmidrule(lr){9-10}
    \cmidrule(lr){11-12}
    \cmidrule(lr){13-14}
    \cmidrule(lr){15-16}
      & & F1 & ACC & F1 & ACC & F1 & ACC & F1 & ACC & F1 & ACC & F1 & ACC & F1 & ACC \\
    \midrule

    \multirow{4}{*}{\shortstack{Image \\ only}}
      & CNNDet\citep{wang2020cnndet}
      & 84.60 & 85.04
      & 71.56 & 75.94
      & 59.70 & 68.72
      & 56.27 & 67.08
      & 35.72 & 57.57
      & 61.57 & 70.87
      & 55.81 & 67.33 \\

      & GramNet\citep{liu2020gramnet}
      & 80.51 & 80.35
      & 74.01 & 76.66
      & 65.28 & 70.76
      & 64.35 & 70.29
      & 52.00 & 63.37
      & 67.23 & 72.29
      & 63.91 & 70.27 \\

      & FreqNet\citep{frank2020freqnet}
      & 75.88 & 77.70
      & 60.97 & 68.37
      & 53.15 & 64.02
      & 53.50 & 64.37
      & 38.47 & 57.08
      & 56.39 & 66.31
      & 51.52 & 63.46 \\

      & NPR\citep{tan2024npr}
      & 79.41 & 79.28
      & 81.67 & 81.84
      & 72.12 & 74.28
      & \underline{73.43} & \underline{75.47}
      & \textbf{67.62} & \textbf{71.36}
      & \underline{74.85} & 76.45
      & \underline{73.71} & 75.74 \\

    \midrule
    
    \multirow{4}{*}{\shortstack{Image \\ \&Pixel}}
      & TruFor\citep{guillaro2023trufor}
      & 90.12 & \textbf{97.73}
      & 35.93 & 55.62
      & 58.04 & 66.41
      & 59.73 & 67.51
      & 49.12 & 61.62
      & 58.59 & 69.78
      & 50.70 & 62.79 \\

      & IML-ViT\citep{ma2023imlvit}
      & \textbf{94.47} & 75.73
      & 69.70 & 61.19
      & 40.98 & 49.95
      & 44.69 & 51.25
      & 18.20 & 43.62
      & 53.61 & 56.35
      & 43.39 & 51.50 \\

      & MaskCLIP\citep{wang2025opensdi}
      & 91.28 & 91.47
      & \underline{83.36} & \underline{85.28}
      & \underline{72.90} & \underline{77.98}
      & 67.79 & 74.95
      & 53.08 & 67.12
      & 73.68 & \underline{79.36}
      & 69.28 & \underline{76.33} \\

      & \textbf{Ours}
      & \underline{93.02} & \underline{93.08}
      & \textbf{91.34} & \textbf{91.73}
      & \textbf{79.89} & \textbf{82.63}
      & \textbf{76.87} & \textbf{80.62}
      & \underline{56.78} & \underline{68.76}
      & \textbf{79.58} & \textbf{83.36}
      & \textbf{76.22} & \textbf{80.93} \\

    \bottomrule
    \vspace{-3em}
  \end{tabular}%
  }
\end{table*}
%%%%%%%%%%%%%%%%%%%%%%%%%%%%%%%%%%%%%%%%%%%%%%%%%%%%%%%%%%%%%%%%%%%%%

%%%%%%%%%%%%%%%%%%%%%%%%%%%%%%%%%%%%%%%%%%%%%%%%%%%%%%%%%%%%%%%%%%%%%
\begin{wraptable}{r}{0.7\columnwidth} 
\vspace{-\baselineskip} 
\centering
\caption{
Ablation studies of ReGFLoW modules.
The module names follow Figure~\ref{fig:fig3}.
}
\label{tab:ablation}
\setlength{\tabcolsep}{0.5pt} 
\renewcommand{\arraystretch}{0.9} 
\small 
\begin{tabular}{lccc|c}
\toprule
Method
& \begin{tabular}{@{}c@{}}Bias \\ Predictor\end{tabular}
& \begin{tabular}{@{}c@{}}Feature-aligned \\ Fusion\end{tabular}
& \begin{tabular}{@{}c@{}}Artifact-Centric \\ MIL\end{tabular}
& Pixel F1 \\
\midrule
ReGFLoW
& \checkmark & \checkmark & \checkmark
& \textbf{49.38} \\

w/o Bias Predictor
$\times$ & & \checkmark & \checkmark
& 46.06 \\

w/o Reconstruction Prior
& $\times$ & $\times$ & \checkmark
& 38.08 \\

Score-Pooling BCE
& $\times$ & $\times$ & $\times$
& 32.34 \\
\bottomrule
\end{tabular}
\end{wraptable}
%%%%%%%%%%%%%%%%%%%%%%%%%%%%%%%%%%%%%%%%%%%%%%%%%%%%%%%%%%%%%%%%%%%%%

\subsection{Ablation Studies}
\label{sec:ablation}
Table~\ref{tab:ablation} ablates the main modules shown in
Figure~\ref{fig:fig3}.
\emph{Feature-aligned fusion} uses the reconstruction prior to modulate local
artifact features through a gated offset, while the \emph{bias predictor}
adds reconstruction-guided evidence directly to the patch score map.
\emph{Artifact-centric MIL} denotes our adaptive top-$k$ aggregation with
fake-real pairwise ranking supervision.

Removing the bias predictor weakens the score-level use of the reconstruction
prior, and removing the reconstruction prior entirely causes a larger drop.
Replacing artifact-centric MIL with image-level BCE on a pooled score further
degrades performance, showing that both reconstruction-guided scoring and
artifact-centric weak supervision are important for localization.

\subsection{Scaling to Broader Training Domains}
\label{sec:broad_domain_training}

%%%%%%%%%%%%%%%%%%%%%%%%%%%%%%%%%%%%%%%%%%%%%%%%%%%%%%%%%%%%%%%%%%%%%
\begin{wraptable}{r}{0.58\columnwidth}
\vspace{-1\intextsep}
\centering
\caption{
Effect of source-domain diversity on ReGFLoW's cross-domain localization. Each row adds one source domain to SD1.5 while keeping the total training size fixed.
}
\label{tab:broad_domain_training}
% \vspace{-0.5em}
\setlength{\tabcolsep}{2.2pt}
\renewcommand{\arraystretch}{0.92}
\resizebox{\linewidth}{!}{%
\begin{tabular}{@{}l|ccccc@{}}
\toprule
Train & SD2 & SD3 & SDXL & FLUX & Avg.\ $\Delta$ \\
\midrule
SD1.5
& 40.96
& 38.09
& 29.36
& 24.12
& -- \\
\midrule
+ SD2
& --
& 42.34 {\tiny (+4.25)}
& 33.73 {\tiny (+4.37)}
& 28.71 {\tiny (+4.60)}
& +4.40 \\

+ SD3
& 42.75 {\tiny (+1.79)}
& --
& 34.48 {\tiny (+5.12)}
& 30.96 {\tiny (+6.84)}
& +4.58 \\

+ SDXL
& 40.88 {\tiny (-0.08)}
& 41.84 {\tiny (+3.75)}
& --
& 31.00 {\tiny (+6.88)}
& +3.52 \\

+ FLUX
& 41.57 {\tiny (+0.61)}
& 43.82 {\tiny (+5.73)}
& 34.47 {\tiny (+5.11)}
& --
& +3.82 \\
\bottomrule
\end{tabular}%
}
\vspace{-0.6\intextsep}
\end{wraptable}
%%%%%%%%%%%%%%%%%%%%%%%%%%%%%%%%%%%%%%%%%%%%%%%%%%%%%%%%%%%%%%%%%%%%%

A practical motivation for weak supervision, as discussed in Section~\ref{sec:rethinking}, is that it enables training on broader and more diverse data sources without requiring pixel-level masks. We test whether ReGFLoW can convert this flexibility into improved cross-domain localization. Using OpenSDID, we extend the SD1.5 training set with one additional source domain while keeping the total number of training images fixed, isolating the effect of domain diversity from data scale by uniformly subsampling each source. As shown in Table~\ref{tab:broad_domain_training}, adding any single source domain consistently improves average pixel-level F1 on unseen domains, with the largest gains observed on the most distant generators where SD1.5-only training provides the weakest baseline. These results show that weak supervision is not merely a lower-cost alternative to dense annotation, but a practical means of scaling diffusion-edited localization to the broader data distributions that pixel-mask supervision cannot accommodate.

\vspace{-0.5em}
\subsection{Robustness to Image Degradations}
\label{sec:robustness}
\vspace{-0.5em}

\begin{figure*}[t]
    \centering
    \includegraphics[width=0.93\textwidth]{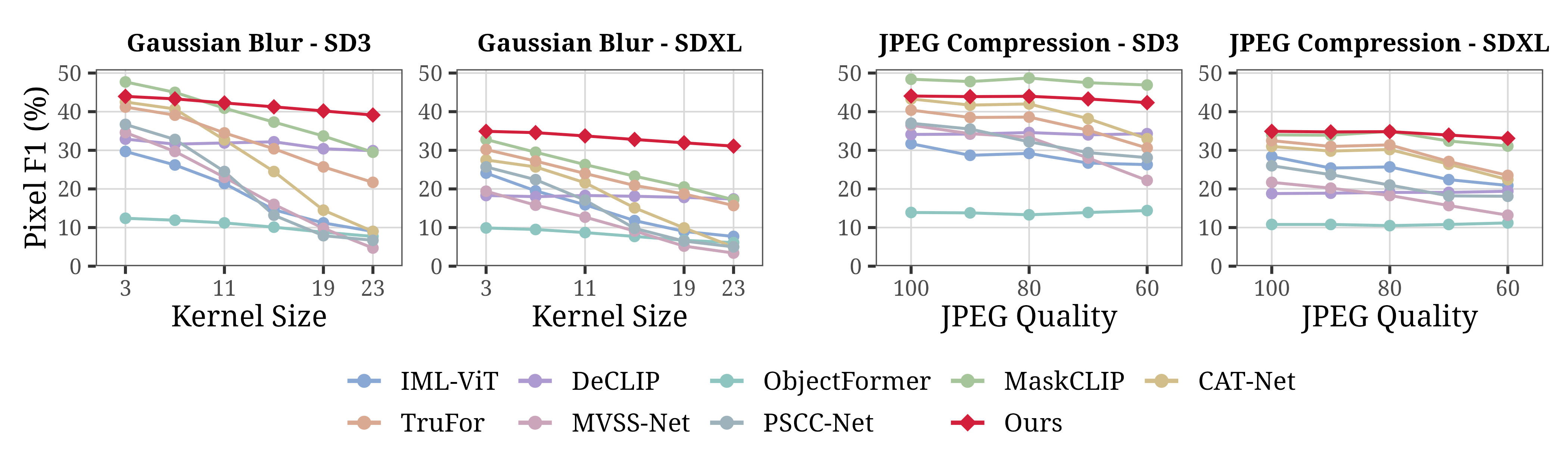}
    \vspace{-0.6em}
    \caption{
    Robustness to Gaussian blur and JPEG compression on SD3 and SDXL.
    Baseline curves are digitized from the robustness results reported in OpenSDI~\citep{wang2025opensdi}, and ReGFLoW is evaluated under the same degradation settings.
    }
    \label{fig:robustness}
    \vspace{-0.8em}
\end{figure*}

We further compare ReGFLoW with additional fully supervised
baselines~\citep{smeu2025declip,wang2022objectformer,chen2021mvssnet,kwon2021catnet} under Gaussian blur and JPEG compression, following the degradation settings reported in~\citep{wang2025opensdi}.
As shown in Figure~\ref{fig:robustness}, ReGFLoW degrades gradually under increasingly strong Gaussian blur and remains stable under JPEG compression across both SD3 and SDXL.
Under severe blur, ReGFLoW outperforms the compared baselines on both generators, while remaining highly competitive under JPEG compression, indicating that its localization is not solely dependent on fragile high-frequency cues.

%%%%%%%%%%%%%%%%%%%%%%%%%%%%%%%%%%%%%%%%%%%%%%%%%%%%%%%%%%%%%%%%%%%%%
%%%%%%%%%%%%%%%%%%%%%%%%%%%%%%%%%%%%%%%%%%%%%%%%%%%%%%%%%%%%%%%%%%%%%

\vspace{-0.5em}
\subsection{Qualitative Analysis}
\label{sec:qualitative}
\begin{figure}[t]
    \centering
    \includegraphics[width=0.85\linewidth]{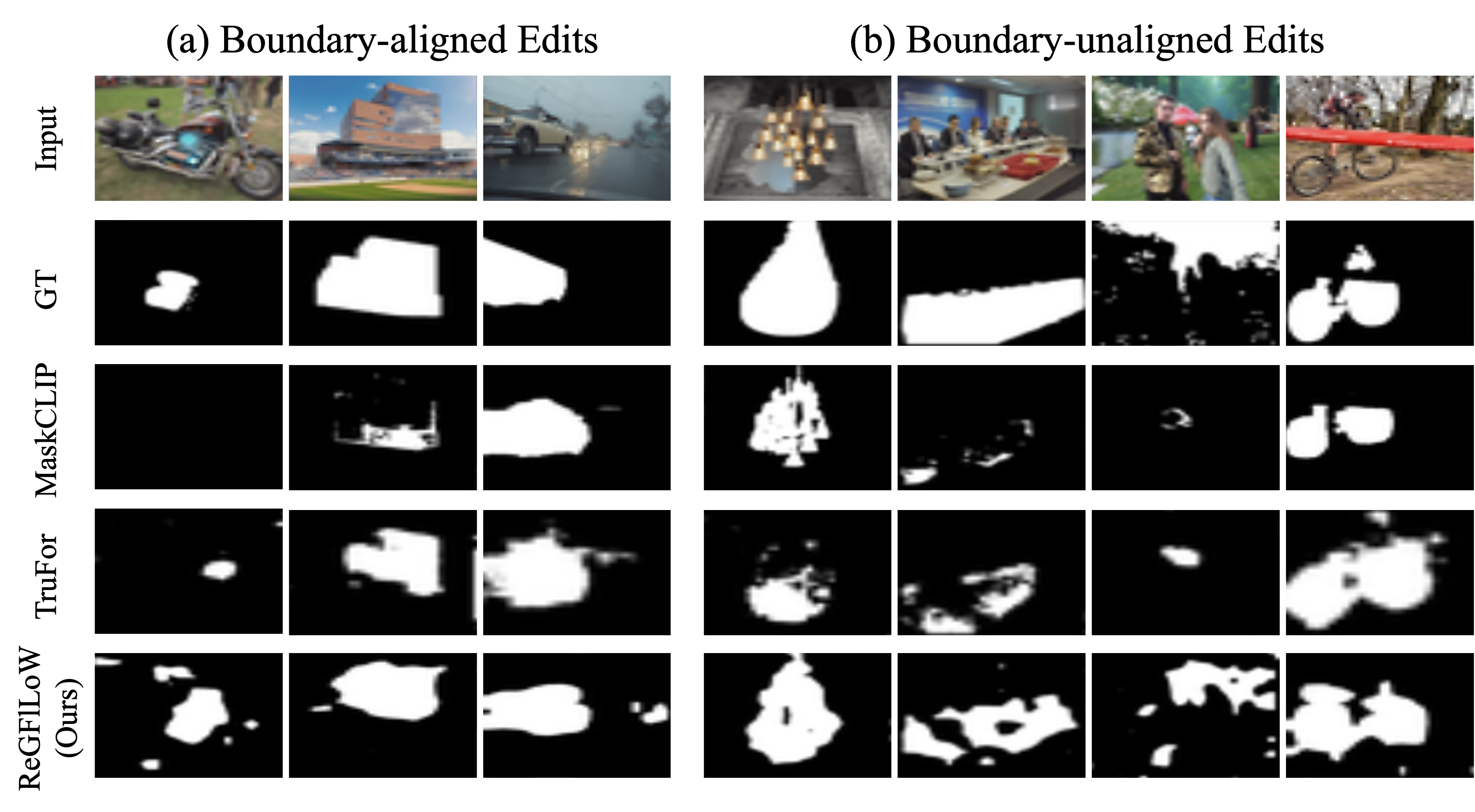}
    \caption{
Qualitative comparison on out-of-domain fake-region localization.
MaskCLIP~\citep{wang2025opensdi} and TruFor~\citep{guillaro2023trufor} are fully supervised, whereas Ours is weakly supervised.
The examples include (a) boundary-aligned edits and (b) boundary-unaligned edits, where boundaries provide limited guidance.
    }
    \vspace{-1em}
    \label{fig:fig4}
\end{figure}

\vspace{-0.5em}
Figure~\ref{fig:fig4} presents qualitative localization results on out-of-domain examples. We include (a) boundary-aligned edits, similar to conventional segmentation-style data, as well as (b) boundary-unaligned cases involving background edits, partial changes inside an object, and multiple manipulated objects. We compare ReGFLoW with fully supervised baselines, MaskCLIP and TruFor. These examples show that when object boundaries provide limited guidance, ReGFLoW better localizes artifact evidence itself rather than simply following semantic or object-aligned regions.

%%%%%%%%%%%%%%%%%%%%%%%%%%%%%%%%%%%%%%%%%%%%%%%%%%%%%%%%%%%%%%%%%%%%%
%%%%%%%%%%%%%%%%%%%%%%%%%%%%%%%%%%%%%%%%%%%%%%%%%%%%%%%%%%%%%%%%%%%%%

\vspace{-0.5em}
\section{Conclusion}
\label{sec:conclusion}
\vspace{-0.5em}
In this paper, we presented ReGFLoW, a weakly-supervised framework for AI-edited fake region localization. 
To address the limitations of fully-supervised methods—namely, the difficulty of collecting scalable annotations and the risk of misleading supervision signals—ReGFLoW learns from image-level real/fake labels and leverages diffusion reconstruction error as dense spatial guidance.
ReGFLoW achieves competitive cross-generator localization and surpasses all evaluated fully supervised baselines when both partially edited and fully synthetic images are considered, as well as in cross-dataset evaluation, without target-domain adaptation.
These results suggest that weakly supervised learning is a practical formulation for learning generalizable features for AI-edited fake region localization as a scalable alternative to mask-supervised frameworks.

%%%%%%%%%%%%%%%%%%%%%%%%%%%%%%%%%%%%%%%%%%%%%%%%%%%%%%%%%%%%%%%%%%%%%
\begin{ack}
This work was supported by the National Research Foundation of Korea (NRF) grant funded by the Korea government(MSIT)(RS-2024-00456589) and Korea Planning \& Evaluation Institute of Industrial Technology(KEIT) grant funded by the Korea government(MOTIE) (No. RS2024-00442120, Development of AI technology capable of robustly recognizing abnormal and dangerous situations and behaviors during night and bad weather conditions)
\end{ack}
%%%%%%%%%%%%%%%%%%%%%%%%%%%%%%%%%%%%%%%%%%%%%%%%%%%%%%%%%%%%%%%%%%%%%
{
\small
\bibliographystyle{unsrtnat}
\bibliography{references}
}
%%%%%%%%%%%%%%%%%%%%%%%%%%%%%%%%%%%%%%%%%%%%%%%%%%%%%%%%%%%%%%%%%%%%%
\newpage
\appendix
% Appendix: requires the main-paper preamble, figures, and bibliography.

\section*{Appendix Overview}
Appendix~\ref{app:mask_availability} discusses mask availability and the
limitations of before--after differencing.
Appendices~\ref{app:loss} and~\ref{app:implementation} detail the training
objective, inference calibration, and implementation.
Appendix~\ref{app:additional_experiments} presents controlled supervision
comparisons, reconstruction-prior analyses, input robustness, and threshold
sensitivity.
Appendices~\ref{app:qualitative}--\ref{app:assets} provide additional
qualitative results, limitations, related work, and asset licenses.

\section{Why Pixel-Level Masks Do Not Scale to Broad Real-World Diffusion Editing}
\label{app:mask_availability}

As discussed in Sec.~\ref{sec:mask_unavailable}, pixel-level masks are
available in controlled, mask-conditioned pipelines but are not a general
supervision source for real-world diffusion editing.
Hosted, conversational, and consumer-facing systems often expose only the
input, instruction, and edited output, not the underlying generator,
editing operation, or full spatial support of generated artifacts.

\subsection{Mask Availability Across Current Image Editing Systems}
\label{app:commercial_mask}

A \emph{conditioning mask} guides an inpainting model; a \emph{UI-level
selection} specifies a brush, click, or region in an editing interface;
and a \emph{fake-region supervision mask} labels the spatial support of
generated or edited artifacts.
The first two describe editing controls, not necessarily the complete
artifact support after latent encoding, denoising, decoding, blending, and
harmonization.

Table~\ref{tab:app_mask_availability} summarizes three workflow categories.
Stable Diffusion/SDXL inpainting, Kandinsky inpainting, and FLUX.1 Fill
explicitly use masks, making them convenient for controlled localization
benchmarks but potentially biasing training toward mask-conditioned sources
\citep{huggingface_diffusers_inpaint,bfl_flux1_tools}.
Consumer-facing tools such as ChatGPT Images, Adobe Firefly, Midjourney
Editor, Canva Magic Edit, and CapCut expose selection-based controls
\citep{openai_chatgpt_images,adobe_firefly_masking,midjourney_editor,canva_magic_edit,capcut_ai_replace}.
These controls may not be preserved when images are shared or collected,
and some APIs operate with or without masks
\citep{openai_image_edit_api,google_imagen_mask_free}.
Instruction- or reference-based models, including Gemini/Nano Banana,
FLUX.2, Seedream, Qwen-Image-Edit, Hunyuan Image, and OmniGen, further
separate editing from explicit spatial annotation
\citep{google_nano_banana,bfl_flux2_docs,bytedance_seedream4,qwen_image_edit,hunyuan_image3,omnigen}.
Their inputs and outputs may be available without reliable dense artifact
labels.

\begin{table}[!htbp]
\centering
\small
\caption{
Mask availability across representative editing workflows.
Edit controls do not necessarily provide reliable pixel-level artifact labels.
}
\label{tab:app_mask_availability}
\begin{tabular}{@{}p{0.27\linewidth} p{0.33\linewidth} p{0.33\linewidth}@{}}
\toprule
\textbf{Workflow category}
&
\textbf{Examples}
&
\textbf{Implication for localization supervision}
\\
\midrule

Mask-conditioned inpainting pipelines
&
Stable Diffusion/SDXL inpainting, Kandinsky inpainting, FLUX.1 Fill
\citep{huggingface_diffusers_inpaint,bfl_flux1_tools}
&
Intended edit masks support controlled benchmarks but may not cover the
full artifact support.
\\

Consumer-facing tools with UI-level selection
&
ChatGPT Images, Adobe Firefly, Midjourney Editor, Canva Magic Edit, CapCut
\citep{openai_chatgpt_images,adobe_firefly_masking,midjourney_editor,canva_magic_edit,capcut_ai_replace}
&
Selections may be lost after sharing and do not reveal the full artifact
support.
\\

Instruction- or reference-based editing models
&
Gemini/Nano Banana, FLUX.2, Seedream, Qwen-Image-Edit, Hunyuan Image, OmniGen
\citep{google_nano_banana,bfl_flux2_docs,bytedance_seedream4,qwen_image_edit,hunyuan_image3,omnigen}
&
Inputs, instructions, references, and outputs may be available without
dense artifact labels.
\\

\bottomrule
\end{tabular}
\end{table}

Image-level real/fake labels therefore support broader training sources
than pixel-level masks, without assuming that all fake evidence is confined
to a known edit region.

\subsection{Why Before--After Image Differences Do Not Provide Reliable Masks}
\label{app:image_diff_mask}

We examine before--after differencing using Gemini, Qwen-Image, and FLUX.2.
Given an original image $I$ and an instruction, we generate an edited image
$\widetilde{I}$, resize it to the original resolution, and compute RGB and
grayscale differences:
\begin{equation}
    \mathbf{D}_{\mathrm{RGB}}
    =
    \left| I - \widetilde{I} \right|,
    \qquad
    \mathbf{D}_{\mathrm{gray}}
    =
    \frac{1}{3}\sum_{c=1}^{3}
    \left| I^{(c)} - \widetilde{I}^{(c)} \right|.
    \label{eq:app_diff_map}
\end{equation}
We use two instructions targeting object replacement and a less localized
scene property, respectively:
\begin{itemize}
    \item \textbf{Object-level edit:}
    ``Replace the red bus in the image with a yellow taxi, keeping the scene
    realistic and consistent.''
    \item \textbf{Scene-level edit:}
    ``Change the scene to daytime by adjusting lighting, shadows, and sky
    naturally.''
\end{itemize}

\begin{figure}[!htbp]
\centering

\includegraphics[width=0.78\linewidth]{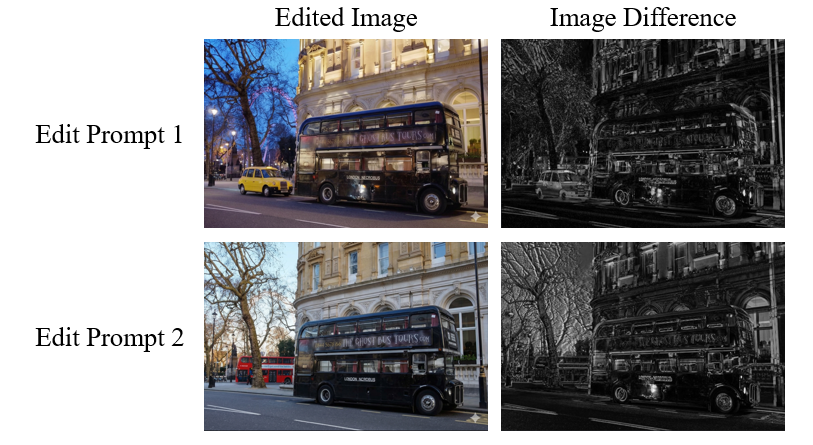}
\vspace{0.5mm}
\centerline{\small (a) Gemini}

\vspace{1.5mm}

\includegraphics[width=0.78\linewidth]{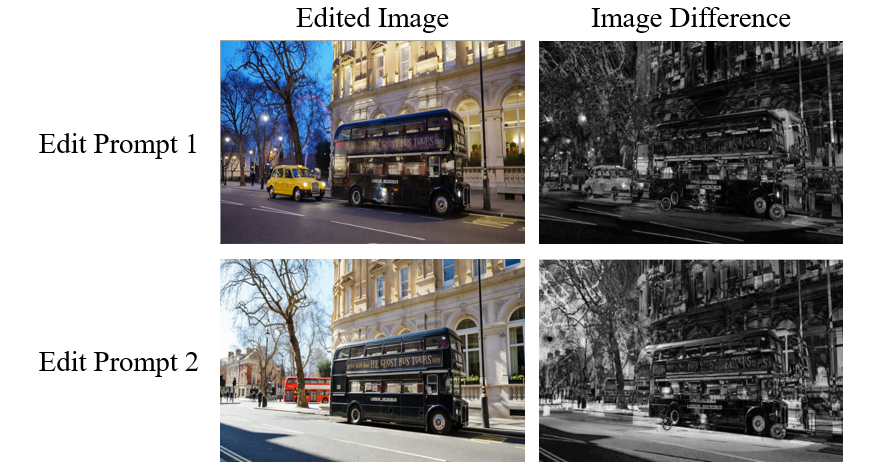}
\vspace{0.5mm}
\centerline{\small (b) Qwen-Image}

\vspace{1.5mm}

\includegraphics[width=0.78\linewidth]{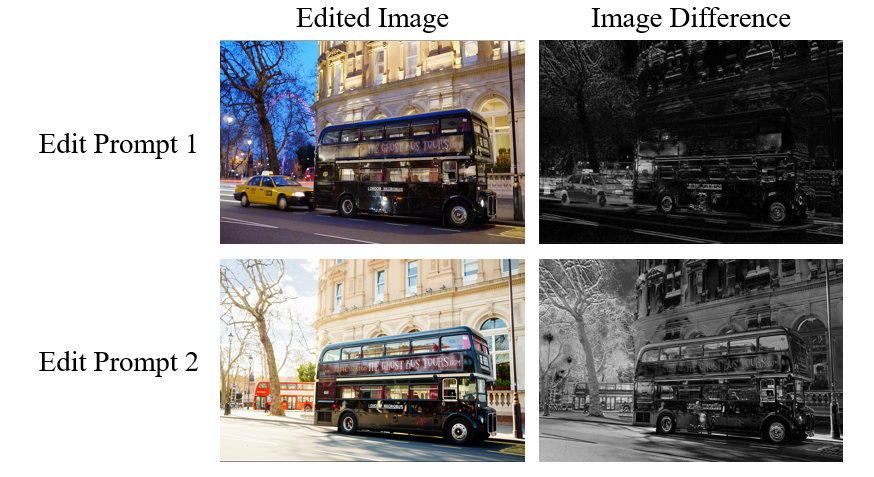}
\vspace{0.5mm}
\centerline{\small (c) FLUX.2}

\caption{
Before--after differencing across three editing systems.
Each panel shows edited outputs and grayscale differences from the original
image. Geometric, photometric, and harmonization changes make these maps
unreliable as pixel-level artifact labels.
}
\label{fig:app_diff_masks}
\end{figure}

Figure~\ref{fig:app_diff_masks} shows that even object-level edits produce
differences outside the intended region through changes in viewpoint,
object scale, boundaries, geometry, texture, color, illumination, shadows,
reflections, and background harmonization.
These are rendering changes rather than simply annotation noise.
Scene-level edits can affect the sky, road, objects, contrast, and color
tone simultaneously, with no single closed target region.
Thresholded difference masks consequently depend on alignment, exposure,
color normalization, and the chosen threshold.

A difference map measures pixel disagreement, not artifact support: it can
respond to geometric or photometric shifts in unchanged semantic regions
while missing artifacts with weak RGB differences.
Global edits, such as lighting changes, background harmonization, or style
modification, further undermine the notion of a single binary edit mask.
Thus, before--after differencing does not resolve the mask-availability
problem and motivates learning artifact evidence from image-level labels.

\section{Additional Details of the Training Objective and Inference Calibration}
\label{app:loss}

All localization losses operate on the MIL logit map
$\mathbf{S}\in\mathbb{R}^{B\times h\times w}$ before upsampling, with
$h=w=16$ by default.
Let $N=hw$ and $\mathbf{s}_i\in\mathbb{R}^{N}$ denote the number of patches
and flattened logits of image $i$.
Labels $y_i=1$ and $y_i=0$ denote fake and real images, whose minibatch
index sets are $\mathcal{F}$ and $\mathcal{R}$.
The full objective is
\begin{equation}
    \begin{aligned}
    \mathcal{L}
    ={}&
    \lambda_{\mathrm{ce}}\,\mathcal{L}_{\mathrm{ce}}
    +
    \lambda_{\mathrm{acmil}}\,\mathcal{L}_{\mathrm{acmil}}
    +
    \lambda_{\mathrm{real}}\,\mathcal{L}_{\mathrm{real}}
    +
    \lambda_{\mathrm{smooth}}\,\mathcal{L}_{\mathrm{smooth}}
    \\
    &+
    \lambda_{\mathrm{sparse}}\,\mathcal{L}_{\mathrm{sparse}}
    +
    \lambda_{\mathrm{con}}\,\mathcal{L}_{\mathrm{con}}
    +
    \lambda_{\mathrm{peak}}\,\mathcal{L}_{\mathrm{peak}}.
    \end{aligned}
    \label{eq:app_full_loss}
\end{equation}
The default weights are
$\lambda_{\mathrm{ce}}=1.0$,
$\lambda_{\mathrm{acmil}}=1.0$,
$\lambda_{\mathrm{real}}=0.05$,
$\lambda_{\mathrm{smooth}}=0.03$,
$\lambda_{\mathrm{sparse}}=3\times10^{-3}$,
$\lambda_{\mathrm{con}}=0.02$, and
$\lambda_{\mathrm{peak}}=0$.
The optional peak-separation term is disabled by default.

\subsection{Real-Image Hard-Negative Suppression}
\label{app:real_loss}

MIL ranking can leave a few high-scoring patches in real images, causing
false positives.
We suppress these responses using the mean of the top fraction
$\rho_{\mathrm{real}}$ of their logits:
\begin{equation}
    T_{\rho}(\mathbf{s}_i)
    =
    \frac{1}{k}
    \sum_{n\in\operatorname{TopK}(\mathbf{s}_i,k)}
    s_{i,n},
    \qquad
    k=\left\lceil \rho N \right\rceil .
    \label{eq:app_topmean}
\end{equation}
\begin{equation}
    \mathcal{L}_{\mathrm{real}}
    =
    \frac{1}{|\mathcal{R}|}
    \sum_{i\in\mathcal{R}}
    \operatorname{softplus}
    \left(
        T_{\rho_{\mathrm{real}}}(\mathbf{s}_i)
    \right).
    \label{eq:app_real_loss}
\end{equation}
This penalizes the most suspicious real-image patches; we use
$\rho_{\mathrm{real}}=0.10$.

\subsection{Edge-Aware Smoothness Regularization}
\label{app:smooth_loss}

To reduce noise without direct boundary supervision, we apply
edge-aware smoothness to fake images in two spatial dimensions rather than
to flattened patch scores.
Let
\begin{equation}
    \mathbf{P}_i = \sigma(\mathbf{S}_i)
    \in [0,1]^{h\times w}
\end{equation}
be the patch-level fake probability map.
We downsample the input image to the same resolution and average its
channels to obtain $\mathbf{G}_i\in\mathbb{R}^{h\times w}$.
Horizontal and vertical differences are
\begin{align}
    \nabla_x \mathbf{P}_{i,u,v}
    &=
    \mathbf{P}_{i,u,v+1}
    -
    \mathbf{P}_{i,u,v},
    \\
    \nabla_y \mathbf{P}_{i,u,v}
    &=
    \mathbf{P}_{i,u+1,v}
    -
    \mathbf{P}_{i,u,v},
\end{align}
with edge-aware weights
\begin{align}
    \mathbf{W}^{x}_{i,u,v}
    &=
    \exp\left(
        -\gamma
        \left|
        \mathbf{G}_{i,u,v+1}
        -
        \mathbf{G}_{i,u,v}
        \right|
    \right),
    \\
    \mathbf{W}^{y}_{i,u,v}
    &=
    \exp\left(
        -\gamma
        \left|
        \mathbf{G}_{i,u+1,v}
        -
        \mathbf{G}_{i,u,v}
        \right|
    \right).
\end{align}
The resulting loss is
\begin{equation}
    \mathcal{L}_{\mathrm{smooth}}
    =
    \frac{1}{|\mathcal{F}|}
    \sum_{i\in\mathcal{F}}
    \left[
    \frac{1}{h(w-1)}
    \sum_{u,v}
    \mathbf{W}^{x}_{i,u,v}
    \left(
        \nabla_x \mathbf{P}_{i,u,v}
    \right)^2
    +
    \frac{1}{(h-1)w}
    \sum_{u,v}
    \mathbf{W}^{y}_{i,u,v}
    \left(
        \nabla_y \mathbf{P}_{i,u,v}
    \right)^2
    \right].
    \label{eq:app_smooth_loss}
\end{equation}
With $\gamma=10.0$, strong image edges reduce smoothing across possible
object or texture boundaries while promoting coherence elsewhere.

\subsection{Fake-Image Sparsity Regularization}
\label{app:sparse_loss}

Because MIL only requires strong fake evidence in some patches, scores may
increase broadly without additional constraints.
We discourage this all-positive solution with
\begin{equation}
    \mathcal{L}_{\mathrm{sparse}}
    =
    \frac{1}{|\mathcal{F}|N}
    \sum_{i\in\mathcal{F}}
    \sum_{n=1}^{N}
    \sigma(s_{i,n}).
    \label{eq:app_sparse_loss}
\end{equation}
Its small weight regularizes the map without overriding MIL, allowing broad
responses when supported by image-level evidence.

\subsection{Patch-Level Contrastive Regularization}
\label{app:contrast_loss}

The MIL head outputs normalized patch embeddings
$\mathbf{E}\in\mathbb{R}^{B\times D\times h\times w}$.
From each image, we select the top $\rho_{\mathrm{con}}$ fraction by patch
logit: selected fake-image patches are artifact-positive, and selected
real-image patches are hard negatives.
Let $\mathcal{A}$ denote this embedding set and $c_a\in\{0,1\}$ its labels,
with $c_a=1$ for fake patches and $c_a=0$ for real patches.
For anchor $a$, the same-class positive set is
\begin{equation}
    \mathcal{P}(a)
    =
    \left\{
        p\in\mathcal{A}
        \;:\;
        p\neq a,\;
        c_p=c_a
    \right\}.
\end{equation}
and the supervised contrastive loss is
\begin{equation}
    \mathcal{L}_{\mathrm{con}}
    =
    -
    \frac{1}{|\mathcal{V}|}
    \sum_{a\in\mathcal{V}}
    \frac{1}{|\mathcal{P}(a)|}
    \sum_{p\in\mathcal{P}(a)}
    \log
    \frac{
        \exp\left(\mathbf{e}_a^\top \mathbf{e}_p / \tau_{\mathrm{con}}\right)
    }{
        \sum_{q\in\mathcal{A},\,q\neq a}
        \exp\left(\mathbf{e}_a^\top \mathbf{e}_q / \tau_{\mathrm{con}}\right)
    },
    \label{eq:app_contrast_loss}
\end{equation}
where $\mathcal{V}$ contains anchors with at least one positive sample.
We use $\rho_{\mathrm{con}}=0.10$ and $\tau_{\mathrm{con}}=0.10$, and set
the loss to zero when a minibatch has insufficient selected fake or real
patches.

\subsection{Optional Peak Separation Regularization}
\label{app:peak_loss}

For a fake image $i$, sort patch logits as
$s_{i,(1)}\geq s_{i,(2)}\geq\cdots\geq s_{i,(N)}$.
The mean top response and mid-ranked background response are
\begin{equation}
    A_i^{\mathrm{top}}
    =
    \frac{1}{k_{\mathrm{top}}}
    \sum_{n=1}^{k_{\mathrm{top}}}
    s_{i,(n)},
    \qquad
    k_{\mathrm{top}}
    =
    \left\lceil
    \rho_{\mathrm{top}}N
    \right\rceil .
\end{equation}
\begin{equation}
    A_i^{\mathrm{rest}}
    =
    \frac{1}{k_{\max}-k_{\min}}
    \sum_{n=k_{\min}}^{k_{\max}}
    s_{i,(n)},
\end{equation}
where $k_{\min}=\lfloor\rho_{\min}N\rfloor$ and
$k_{\max}=\lceil\rho_{\max}N\rceil$.
The optional loss separates these responses by a margin:
\begin{equation}
    \mathcal{L}_{\mathrm{peak}}
    =
    \frac{1}{|\mathcal{F}|}
    \sum_{i\in\mathcal{F}}
    \max
    \left(
        0,\;
        m_{\mathrm{peak}}
        -
        \left(
            A_i^{\mathrm{top}}
            -
            A_i^{\mathrm{rest}}
        \right)
    \right).
    \label{eq:app_peak_loss}
\end{equation}
This term is implemented for ablation but disabled in the default
configuration ($\lambda_{\mathrm{peak}}=0$).

\subsection{Image-Wise Adaptive Calibration}
\label{app:calib}

Image-level supervision permits the absolute logit scale to vary across
images.
At inference only, we calibrate each prediction map using an adaptive
raw-logit offset, then apply a fixed probability threshold.
Let $\mathbf{U}_i\in\mathbb{R}^{H\times W}$ be
$\mathbf{S}_i$ bilinearly upsampled to the input resolution.
Its uncalibrated probabilities and soft positive-response ratio are
\begin{equation}
    \mathbf{P}^{0}_i
    =
    \sigma(\mathbf{U}_i),
\end{equation}
\begin{equation}
    r_i
    =
    \frac{1}{HW}
    \sum_{u=1}^{H}
    \sum_{v=1}^{W}
    \mathbf{P}^{0}_{i,u,v}.
    \label{eq:app_raw_ratio}
\end{equation}
Here, $r_i$ averages continuous probabilities rather than counting pixels
above a binary threshold.
The raw-logit statistics are
\begin{equation}
    \mu_i
    =
    \frac{1}{HW}
    \sum_{u,v}
    \mathbf{U}_{i,u,v},
    \qquad
    \sigma_i
    =
    \sqrt{
    \frac{1}{HW}
    \sum_{u,v}
    \left(
        \mathbf{U}_{i,u,v}
        -
        \mu_i
    \right)^2
    }.
    \label{eq:app_logit_stats}
\end{equation}
Using a fixed reference response $c$, shared across images and domains, we
compute
\begin{equation}
    d_i
    =
    \operatorname{clip}
    \left(
        \frac{r_i-c}{\max(c,\epsilon)},
        -1,
        1
    \right),
    \label{eq:app_ratio_delta}
\end{equation}
\begin{equation}
    k_i
    =
    \operatorname{clip}
    \left(
        k_0
        -
        g d_i,
        k_{\min},
        k_{\max}
    \right),
    \label{eq:app_adaptive_k}
\end{equation}
where $k_0$ is the base coefficient and $g$ controls its adjustment.
The adaptive offset and calibrated maps are
\begin{equation}
    \tau_i
    =
    \mu_i
    +
    k_i\sigma_i.
    \label{eq:app_tau}
\end{equation}
\begin{equation}
    \widetilde{\mathbf{U}}_i
    =
    s_{\mathrm{calib}}
    \frac{
        \mathbf{U}_i-\tau_i
    }{
        \sigma_i+\epsilon
    },
    \qquad
    \widehat{\mathbf{M}}_i
    =
    \sigma
    \left(
        \widetilde{\mathbf{U}}_i
    \right).
    \label{eq:app_calib}
\end{equation}
We threshold $\widehat{\mathbf{M}}_i$ at $t=0.5$.
Thus, $\tau_i$ is an image-dependent centering offset in raw-logit space,
whereas $t$ is a global threshold in calibrated probability space; at
$t=0.5$, the raw-logit decision boundary is $\tau_i$.

The default parameters are $k_0=0.9$, $s_{\mathrm{calib}}=2.0$,
$\epsilon=10^{-6}$, $c=0.08$, $g=0.50$, $k_{\min}=0.40$, and
$k_{\max}=1.20$.
Calibration uses only the current image's prediction map, with no
pixel-level ground truth, batch- or dataset-level statistics, threshold
search, or target-domain tuning.
The rule, parameters, and default threshold remain unchanged across
in-domain, out-of-domain, and cross-dataset evaluations.
Appendix~\ref{app:threshold_sensitivity} varies only the final threshold
to assess sensitivity.

\section{Implementation Details}
\label{app:implementation}

ReGFLoW is implemented as a model module in IMDLBenCo, following the
OpenSDI protocol and using the framework's data-loading, distributed-training,
logging, and evaluation utilities.

\paragraph{Backbone configuration.}
The global encoder is a frozen CLIP ViT-L/14~\citep{radford2021clip}.
The trainable local encoder follows the SideAdapterNetwork design in
OpenSDI/IMDLBenCo and uses an ImageNet-pretrained MAE ViT-B/32
backbone~\citep{he2022mae}.
Task-specific heads are trainable.
Inputs are resized to $512\times512$ for the local branch and to the
standard CLIP resolution for the global branch.

\paragraph{Training setup.}
Unless otherwise specified, training uses the OpenSDID-SD1.5 split with
image-level real/fake labels; pixel-level masks are reserved for evaluation.
We train ReGFLoW for 10 epochs on a single GPU with a batch size of 8.
AdamW uses a learning rate of $5\times10^{-5}$, weight decay of $0.05$,
and no warmup; the random seed is 42.
Augmentations include random scaling, flips, $90^\circ$ rotations, mild
brightness/contrast changes, image compression, and Gaussian blur.

\paragraph{Reconstruction prior.}
Each sample includes a single-channel $128\times128$ residual map for the
reconstruction-guided modules.
The map is resized as needed and converted to the input tensor type.

\paragraph{Optimization hyperparameters.}
Local-feature pooling gives a $16\times16$ MIL scoring grid.
The scoring head has a hidden dimension of 256, adaptive MIL pooling uses
sharpness $\kappa=2.0$, and the ranking margin is $0.5$.
Loss definitions, weights, and hard-negative selection are specified in
Appendix~\ref{app:loss}.
New heads use a learning-rate multiplier of $3.0$.

\paragraph{Evaluation.}
Pixel-level F1 and IoU are computed from the calibrated localization map
using the inference procedure in Appendix~\ref{app:calib}.
Image-level F1 and accuracy use the image classifier output.

\section{Additional Quantitative Analyses}
\label{app:additional_experiments}

We report pixel-level F1 (\%) on OpenSDID.
SD1.5 is the training domain; SD2.1, SDXL, SD3, and Flux.1 are unseen
generators, whose unweighted mean is OOD Avg.
The supervision comparisons use P+F, including partially edited and fully
synthetic images during training and evaluation.
The remaining analyses use P, evaluating partially edited images.

\subsection{Effect of Supervision under Matched Backbones}
\label{app:supervision_controls}

To distinguish supervision-related effects from differences between
independently designed architectures, we retain each backbone architecture
and its non-head components while changing the localization head and
supervision.

\paragraph{MaskCLIP backbone.}
MaskCLIP-WS replaces MaskCLIP's fully supervised segmentation head with our
MIL head trained using image-level labels, retaining its backbone and
non-head components~\citep{wang2025opensdi}.

\paragraph{ReGFLoW backbone.}
ReGFLoW-FS retains the ReGFLoW backbone and reconstruction guidance but
replaces its MIL head with a segmentation head trained using pixel-level
masks.
All variants are trained on OpenSDID-SD1.5 under the same P+F protocol.

\begin{table}[htbp]
\centering
\small
\setlength{\tabcolsep}{4pt}
\caption{
Controlled supervision comparisons under P+F.
Each pair shares the backbone architecture and non-head components but
changes the localization head and supervision.
Pixel-level F1 (\%); bold marks the better result within each pair.
}
\label{tab:app_supervision_controls}
\begin{tabular}{llrrrrrr}
\toprule
Supervision & Method & SD1.5 & SD2.1 & SDXL & SD3 & Flux.1 & OOD Avg. \\
\midrule
\multicolumn{8}{l}{\textit{(a) MaskCLIP backbone}} \\
Full & MaskCLIP
& \textbf{95.1} & \textbf{73.2} & 20.8 & 17.7 & 4.8 & \textbf{29.1} \\
Weak & MaskCLIP-WS
& 43.6 & 34.9 & \textbf{28.9} & \textbf{29.4} & \textbf{20.7} & 28.5 \\
\midrule
\multicolumn{8}{l}{\textit{(b) ReGFLoW backbone with reconstruction guidance}} \\
Full & ReGFLoW-FS
& \textbf{92.22} & \textbf{65.26} & 16.89 & 15.46 & 3.41 & 25.25 \\
Weak & ReGFLoW
& 35.6 & 33.4 & \textbf{30.1} & \textbf{32.2} & \textbf{29.4} & \textbf{31.3} \\
\bottomrule
\end{tabular}
\end{table}

Table~\ref{tab:app_supervision_controls} shows the same pattern for both
backbones: full supervision performs better on SD1.5 and SD2.1, whereas
weak supervision improves SDXL, SD3, and Flux.1.
ReGFLoW increases OOD Avg. from 25.25 to 31.3 relative to ReGFLoW-FS.
MaskCLIP-WS has a comparable but slightly lower OOD Avg. than MaskCLIP
(28.5 versus 29.1), reflecting the fully supervised model's SD2.1 advantage.
These results support a contribution from the weakly supervised
formulation beyond backbone choice or reconstruction guidance, but not a
uniform gain across domains.
Because the head and supervision change together, the comparisons assess
the learning formulation rather than label granularity alone.

\subsection{Reconstruction-Residual Localization Baseline}
\label{app:residual_baseline}

DIRE~\citep{wang2023dire} and AEROBLADE~\citep{ricker2024aeroblade}
primarily address image-level detection; AEROBLADE also demonstrates
qualitative localization of inpainted regions.
To test our prior's standalone localization ability, we apply ReGFLoW's
per-image normalization and final thresholding procedure ($t=0.5$) directly
to the SD1-VAE residual map, without training or dataset-specific tuning.
This controlled baseline is not a reproduction of either complete method.

\begin{table}[htbp]
\centering
\small
\setlength{\tabcolsep}{7pt}
\caption{
Residual-only localization versus ReGFLoW under P, using the same per-image
normalization and final thresholding procedure.
Values are pixel-level F1 (\%).
}
\label{tab:app_residual_baseline}
\begin{tabular}{lrrrrr}
\toprule
Method & SD1.5 & SD2.1 & SDXL & SD3 & Flux.1 \\
\midrule
Recon. residual & 18.5 & 22.1 & 17.4 & 18.5 & 13.6 \\
ReGFLoW
& \textbf{49.3} & \textbf{46.3} & \textbf{34.9} & \textbf{43.9} & \textbf{29.2} \\
\bottomrule
\end{tabular}
\end{table}

ReGFLoW exceeds residual thresholding by 15.6--30.8 F1 points across the
five generators (Table~\ref{tab:app_residual_baseline}).
Under this inference rule, the residual is substantially weaker as a final
localization map than as auxiliary guidance learned jointly with RGB
evidence and artifact-centric MIL.
The main-paper ablations examine the individual module contributions.

\subsection{Robustness to Alternative Reconstruction Priors}
\label{app:prior_robustness}

We replace only the inference-time prior with maps from DIRE-DDIM
reconstruction~\citep{wang2023dire} or an SDXL
VAE~\citep{podell2024sdxl}.
The RGB input, SD1-VAE-trained checkpoint, learned parameters, and inference
rule remain fixed, with no retraining, fine-tuning, adaptation, or
prior-specific recalibration.
This tests train--test mismatch in the auxiliary prior.

\begin{table}[htbp]
\centering
\small
\setlength{\tabcolsep}{6pt}
\caption{
Inference-time prior replacement under P, with the RGB input and learned
parameters unchanged. Values are pixel-level F1 (\%).
}
\label{tab:app_prior_robustness}
\begin{tabular}{lrrrrr}
\toprule
Reconstruction prior & SD1.5 & SD2.1 & SDXL & SD3 & Flux.1 \\
\midrule
DIRE-DDIM & 46.7 & 43.7 & 32.7 & 40.8 & 25.9 \\
SDXL VAE & 47.0 & 43.4 & 33.5 & 41.3 & 26.2 \\
SD1 VAE (default)
& \textbf{49.3} & \textbf{46.3} & \textbf{34.9} & \textbf{43.9} & \textbf{29.2} \\
\bottomrule
\end{tabular}
\end{table}

The alternative priors reduce F1 by 1.4--3.3 points
(Table~\ref{tab:app_prior_robustness}).
Although the default performs best, both replacements retain much of the
localization performance, indicating tolerance to the tested alternatives
rather than exclusive dependence on the training reconstructor.

\subsection{Robustness to Resizing and Input Reconstruction}
\label{app:input_robustness}

We extend the main-paper blur and JPEG evaluations to $0.5\times$ resizing
and \emph{edit-then-reconstruct}, which reconstructs the edited input with
an SDXL VAE before inference.
The checkpoint and inference parameters remain fixed, without retraining
or transformation-specific tuning.
Unlike prior replacement in Appendix~\ref{app:prior_robustness},
edit-then-reconstruct changes the observed image itself.

\begin{table}[htbp]
\centering
\small
\setlength{\tabcolsep}{6pt}
\caption{
Input-transformation robustness under P.
Edit-then-reconstruct applies an SDXL VAE to the edited image before inference.
Values are pixel-level F1 (\%).
}
\label{tab:app_input_robustness}
\begin{tabular}{lrrrrr}
\toprule
Input condition & SD1.5 & SD2.1 & SDXL & SD3 & Flux.1 \\
\midrule
Clean & 49.34 & 46.27 & 34.86 & 43.90 & 29.18 \\
Resize ($0.5\times$) & 48.8 & 45.5 & 34.5 & 43.3 & 28.7 \\
Edit-then-reconstruct & 44.8 & 42.1 & 30.3 & 39.6 & 26.6 \\
\bottomrule
\end{tabular}
\end{table}

Resizing reduces F1 by less than 0.8 points per generator, while
edit-then-reconstruct reduces it by approximately 2.6--4.6 points
(Table~\ref{tab:app_input_robustness}).
Together with the main-paper results, these tests cover compression,
spatial smoothing, resizing, and reconstruction-based post-processing,
complementing the separate auxiliary-prior mismatch analysis.

\subsection{Sensitivity to the Final Decision Threshold}
\label{app:threshold_sensitivity}

We retain the checkpoint and image-wise calibration in
Appendix~\ref{app:calib} and vary only the final probability threshold $t$.
This is a diagnostic sweep, not domain-specific threshold selection;
$t=0.5$ remains the shared default for all evaluations.

\begin{table}[htbp]
\centering
\small
\setlength{\tabcolsep}{6pt}
\caption{
Final-threshold sensitivity under P, with the checkpoint and calibration
fixed. Values are pixel-level F1 (\%); bold marks the best result per
generator, including ties. The shared default is $t=0.5$.
}
\label{tab:app_threshold_sensitivity}
\begin{tabular}{lrrrrr}
\toprule
Final threshold $t$ & SD1.5 & SD2.1 & SDXL & SD3 & Flux.1 \\
\midrule
0.4 & 48.20 & 45.68 & \textbf{35.82} & 43.42 & \textbf{30.18} \\
0.5 (default) & \textbf{49.34} & \textbf{46.27} & 34.86 & \textbf{43.90} & 29.18 \\
0.6 & \textbf{49.34} & 45.63 & 32.78 & 43.17 & 27.26 \\
0.7 & 47.75 & 43.33 & 29.21 & 40.65 & 24.06 \\
0.8 & 43.40 & 38.01 & 23.07 & 34.97 & 18.86 \\
\bottomrule
\end{tabular}
\end{table}

Performance is relatively stable over $t\in[0.4,0.6]$, with at most
3.04 F1 points of within-generator variation
(Table~\ref{tab:app_threshold_sensitivity}).
Higher thresholds consistently reduce F1 as masks become more conservative.
SDXL and Flux.1 favor $t=0.4$, whereas SD1.5, SD2.1, and SD3 favor
$0.5$ or $0.6$, confirming that the shared default does not maximize every
domain's score.

\section{Additional Qualitative Results}
\label{app:qualitative}

\subsection{Qualitative Analysis}
\label{sec:qualitative_appendix}

\begin{figure}[!htbp]
\centering
\includegraphics[width=\linewidth]{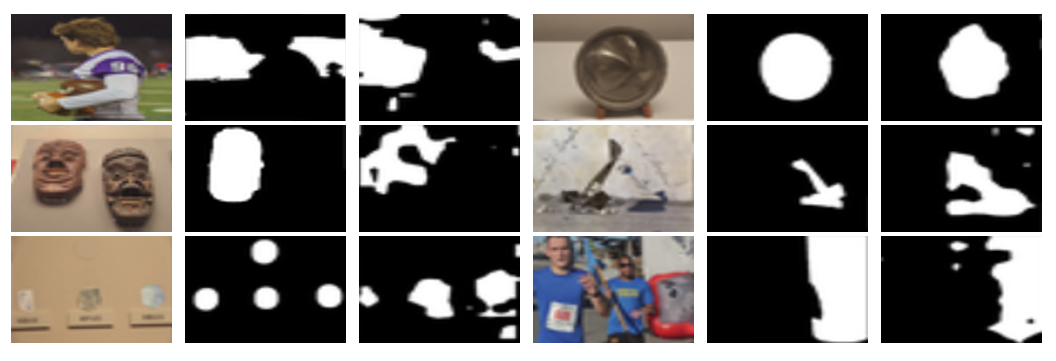}
\vspace{1mm}

\includegraphics[width=\linewidth]{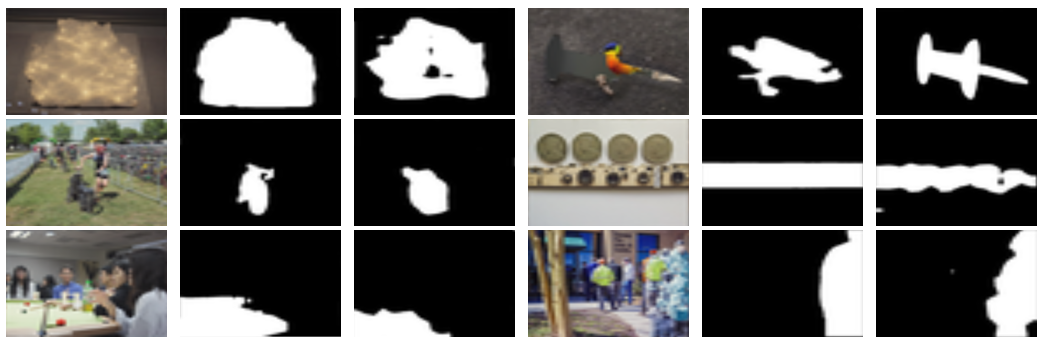}
\vspace{1mm}

\includegraphics[width=\linewidth]{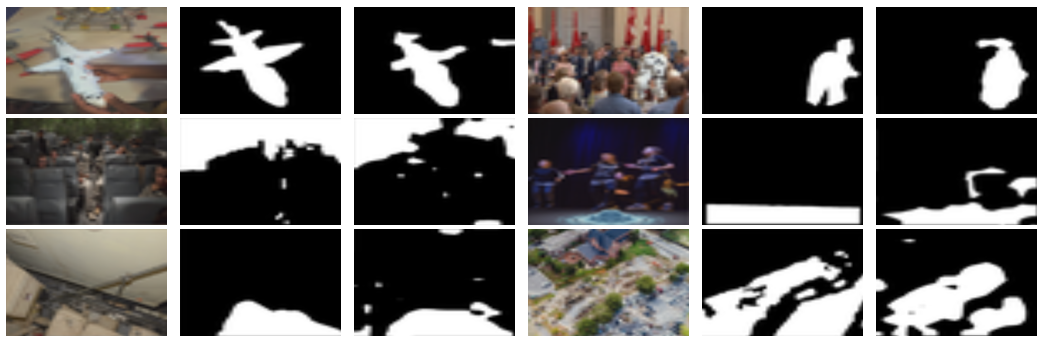}

\caption{
Additional out-of-domain localization results.
Each triplet shows the input image, ground-truth mask, and ReGFLoW prediction
from left to right.
}
\label{fig:app_qualitative}
\end{figure}

Figure~\ref{fig:app_qualitative} extends the main-paper examples to
mask-confined edits, boundary or out-of-mask traces, and non-semantic
background changes.
The predictions illustrate responses to broader artifact support rather
than only object-aligned regions.
Real-image false-positive controls and difficult cases involving tiny edits,
compression artifacts, or high-frequency real textures illustrate the
remaining limitations under cross-domain shifts.

\section{Limitations}
\label{app:limitations}

ReGFLoW requires an additional reconstruction step to obtain its residual
prior, increasing preprocessing or inference cost over RGB-only methods
despite using a frozen autoencoder.
Patch-level MIL can also miss edits smaller than or comparable to the patch
resolution, or activate neighboring pixels, reducing localization precision
and F1 for small, sharply defined masks.
Finally, the controlled comparisons in
Appendix~\ref{app:supervision_controls} reveal a source-domain/transfer
trade-off: full supervision remains stronger on SD1.5 and SD2.1.
The observed benefits therefore do not imply uniformly better performance
on every domain.

\section{Related Work}
\label{app:related_work}

\subsection{From Image-Level AI-Generated Image Detection to Pixel-Level Localization}
\label{app:rw_image_to_pixel}

DIRE uses diffusion reconstruction error for image-level detection, based on
the observation that generated images can be reconstructed more accurately
than real images~\citep{wang2023dire}.
AEROBLADE uses autoencoder reconstruction error for training-free detection
and demonstrates qualitative localization of inpainted
regions~\citep{ricker2024aeroblade}.
RINE exploits intermediate CLIP representations rather than only final
semantic features~\citep{koutlis2024rine}.
These methods motivate reconstruction behavior and pretrained features as
forensic evidence, but an image-level authenticity score alone does not
identify a small or spatially diffuse edited region.
Dense localization requires spatial predictions; the residual-only
comparison in Appendix~\ref{app:residual_baseline} tests our prior under
this protocol.

\subsection{From Conventional Image Manipulation Localization to Diffusion-Edited Localization}
\label{app:rw_iml_to_diffusion}

Conventional localization addresses splicing, copy-move, removal, and
inpainting.
IML-ViT emphasizes high-resolution processing, multi-scale features, and
manipulation-edge supervision~\citep{ma2023imlvit}.
TruFor fuses RGB and learned noise-sensitive fingerprints with a transformer
to predict localization, integrity, and reliability~\citep{guillaro2023trufor}.
HiFi-IFDL learns hierarchical forgery attributes across manipulation types
and generation sources for detection and localization~\citep{guo2023hifi}.

Such methods often exploit boundaries, high-frequency inconsistencies,
camera/noise patterns, or differences between authentic and manipulated
regions.
Diffusion editing can weaken these cues by blending or harmonizing nearby
content, while object replacement, background generation, texture edits,
and lighting changes need not have clear boundaries.
TGIF evaluates this transition with text-guided inpainting
forgeries~\citep{mareen2024tgif}: conventional localizers can detect
spliced regions but struggle with fully regenerated images, whereas
synthetic-image detectors recognize regeneration without localizing the
inpainted region.

\subsection{Diffusion-Era Localization Benchmarks and the Need for Weak Supervision}
\label{app:rw_diffusion_benchmarks}

X-Edit localizes text-guided edits using diffusion-inversion features and
paired original/edited images with masks~\citep{bazyleva2025xedit}.
DEAL-300K combines instruction generation, mask-free editing, and an
annotation pipeline to obtain pixel-level labels~\citep{zhang2025deal300k}.
SIDA provides a multimodal detection, localization, and explanation
framework with annotated social-media synthetic and tampered
images~\citep{huang2025sida}.
TGIF2 extends inpainting evaluation to newer generators, including FLUX.1,
and random non-semantic masks that probe object-centric
biases~\citep{mareen2026tgif2}.
BR-Gen targets underrepresented background regions such as sky, ground,
wall, grass, and vegetation~\citep{cai2026brgen}.
RADAR introduces BBC-PAIR, containing manipulations from 28 diffusion
models, and combines semantic and geometric foundation-model features with
contrastive learning~\citep{costanzino2025radar}.
OpenSDI studies open-world detection and localization through MaskCLIP,
aligning CLIP with MAE for globally and locally manipulated
images~\citep{wang2025opensdi}.
Despite these advances, reliance on paired images, dense masks, or
large-scale annotation restricts training to sources where such
supervision is available.

Weak supervision reduces this dependence.
WSCL learns from image-level labels through multi-source and inter-patch
consistency, using RGB, SRM, and Bayar streams for generic
manipulations~\citep{zhai2023wscl}.
DOLOS compares explanation-, attention-, and local-score-based approaches
for weakly supervised localization of diffusion-generated faces under
controlled generator settings~\citep{tantaru2024dolos}.

ReGFLoW learns from image-level labels for general diffusion-generated and
diffusion-edited images.
It combines dense reconstruction guidance with artifact-centric MIL instead
of object-centric pseudo masks, manipulation boundaries, or camera/noise
priors.

\section{Existing Assets and Licenses}
\label{app:assets}

Table~\ref{tab:assets_licenses} lists the datasets, codebases, pretrained
models, and libraries used for research, with their original citations and
available license terms.

\begin{table}[!htbp]
\centering
\small
\caption{
Existing assets used in this work.
}
\label{tab:assets_licenses}
\begin{tabular}{p{0.24\linewidth} p{0.39\linewidth} p{0.27\linewidth}}
\toprule
\textbf{Asset}
&
\textbf{Usage in this work}
&
\textbf{License / terms}
\\
\midrule

OpenSDID / OpenSDI~\citep{wang2025opensdi}
&
Training and evaluation dataset for diffusion-generated and diffusion-edited
image detection/localization.
&
CC BY-SA 4.0; academic use.
\\

IMDLBenCo~\citep{ma2025imdl}
&
Training, evaluation, logging, and metric computation framework.
&
CC BY 4.0.
\\

CLIP~\citep{radford2021clip}
&
Frozen global visual encoder.
&
MIT License.
\\

MAE~\citep{he2022mae}
&
Initialization for the local artifact encoder.
&
CC BY-NC 4.0.
\\

Stable Diffusion VAE / latent diffusion model~\citep{rombach2022ldm}
&
Frozen reconstruction model used to compute the reconstruction residual prior.
&
CreativeML OpenRAIL-M.
\\

Hugging Face Diffusers
&
Software library used for loading or running diffusion/VAE components.
&
Apache License 2.0.
\\

\bottomrule
\end{tabular}
\end{table}

%%%%%%%%%%%%%%%%%%%%%%%%%%%%%%%%%%%%%%%%%%%%%%%%%%%%%%%%%%%%%%%%%%%%%

\end{document}